\documentclass[10pt,twocolumn,twoside]{IEEEtran}

\pdfoutput=1
\usepackage[utf8]{inputenc} 
\usepackage[T1]{fontenc}    
\usepackage{hyperref}       
\usepackage{url}            
\usepackage{booktabs}       
\usepackage{amsfonts}       
\usepackage{nicefrac}       
\usepackage{microtype}      

\usepackage[pdftex]{graphicx}
\DeclareGraphicsExtensions{.pdf,.jpeg,.png}
\usepackage[cmex10]{amsmath}
\usepackage{amssymb}
\usepackage{amsthm}
\usepackage{amsfonts}
\usepackage{epstopdf}
\usepackage{multirow}
\usepackage{algorithm}
\usepackage{algorithmic}
\usepackage{subfigure}
\usepackage{url}
\usepackage{color}
\usepackage{mathtools}

\newcommand{\Rbb}{\mathbb{R}}

\newcommand{\E}{{\mathcal{E}}}

\newcommand{\D}{{\mathcal{D}}}
\newcommand{\V}{{\mathcal{V}}}

\renewcommand{\L}{{\mathcal{L}}}

\usepackage{mathtools}

\newtheorem{proposition}{{Proposition}}

\DeclareMathOperator{\prox}{prox}

\DeclareMathOperator{\tr}{tr}

\begin{document}
\title{Learning heat diffusion graphs}

\author{Dorina~Thanou, 
          Xiaowen~Dong, 
          Daniel~Kressner,
        and Pascal~Frossard
\thanks{D. Thanou, and P. Frossard  are with the Signal Processing Laboratory (LTS4), \'{E}cole Polytechnique F\'{e}d\'{e}rale de Lausanne (EPFL), Lausanne, Switzerland (e-mail: \{dorina.thanou, pascal.frossard\}@epfl.ch).}
\thanks{X. Dong is with the MIT Media Lab, Cambridge, MA, United States (e-mail: xdong@mit.edu).}
\thanks{D. Kressner is with the Chair of Numerical Algorithms and High-Performance Computing (ANCHP), \'{E}cole Polytechnique F\'{e}d\'{e}rale de Lausanne (EPFL), Lausanne, Switzerland (e-mail: daniel.kressner@epfl.ch). }}

\maketitle

\begin{abstract}
Effective information analysis generally boils down to properly identifying the structure or geometry of the data, which is often represented by a graph. In some applications, this structure may be partly determined by design constraints or pre-determined sensing arrangements, like in road transportation networks for example. In general though, the data structure is not readily available and becomes pretty difficult to define. In particular, the global smoothness assumptions, that most of the existing works adopt,  are often too general and unable to properly capture localized properties of data. In this paper, we go beyond this classical data model and rather propose to represent information as a sparse combination of localized functions that live on a data structure represented by a graph. Based on this model, we focus on the problem of inferring the connectivity that best explains the data samples at different vertices of a graph that is a priori unknown. We concentrate on the case where the observed data is actually the sum of heat diffusion processes, which is a quite common model for data on networks or other irregular structures. We cast a new graph learning problem and solve it with an efficient nonconvex optimization algorithm. Experiments on both synthetic and real world data finally illustrate the benefits of the proposed graph learning framework and confirm that the data structure can be efficiently learned from data observations only.
We believe that our algorithm will help solving key questions in diverse application domains such as social and biological network analysis 
where it is crucial to unveil proper geometry for data understanding and inference.
\end{abstract}

\IEEEpeerreviewmaketitle

\section{Introduction}

Data analysis and processing tasks typically involve large sets of structured data, where the structure carries critical information about the nature of the recorded signals. One can find numerous examples of such datasets in a wide diversity of application domains, such as transportation networks, social or computer networks, brain analysis or even digital imaging and vision. Graphs are commonly used to describe the structure or geometry of such data as they provide a flexible tool for representing and eventually manipulating information that resides on topologically complicated domains.  Once an appropriate graph is constructed, inference and analysis tasks can be carried out with a careful consideration of the data geometry using, e.g., spectral theory~\cite{Chung97} or graph signal processing~\cite{Shuman13a} concepts. The most common model that relates data and graph topology consists in assuming that the data is globally smooth on the graph, that is, samples connected by strong edge weights tend to be similar.  This global model is, however, not always capable of properly capturing specific properties of the data.  While recent research has put a focus on the development of effective methods for processing data on graphs and networks, relatively little attention has been given to the definition of good graphs. This problem remains critical and may actually represent the major obstacle towards effective processing of structured data.

In this work, we first depart from the common globally smooth signal model and  propose a more generic model where the data consists of (sparse) combinations of overlapping local patterns that reside on the graph. These patterns may describe localized events or specific processes appearing at different vertices of the graph, such as traffic bottlenecks in transportation networks or rumor sources in social networks.
More specifically, we view the data measurements as  observations at different time instants of a few processes that start at different nodes of an unknown graph and diffuse with time. Such data can be represented as the combination of graph heat kernels or, more generally, of localized graph kernels.  Particularly the heat diffusion model can be widely applied in real world scenarios to understand the distribution of heat (sources) \cite{Chung11122007}. 
One example is the propagation of a heat wave in geographical spaces.  Another example is the movement of people in buildings or vehicles in cities, which are represented on a geographical graph. 
Finally, a shift of people's interest towards certain subjects on social media platforms such as Twitter could also be understood via a heat diffusion model \cite{Ma_2008}. 

\begin{figure*}
      \centering
            { \includegraphics[width=18cm]{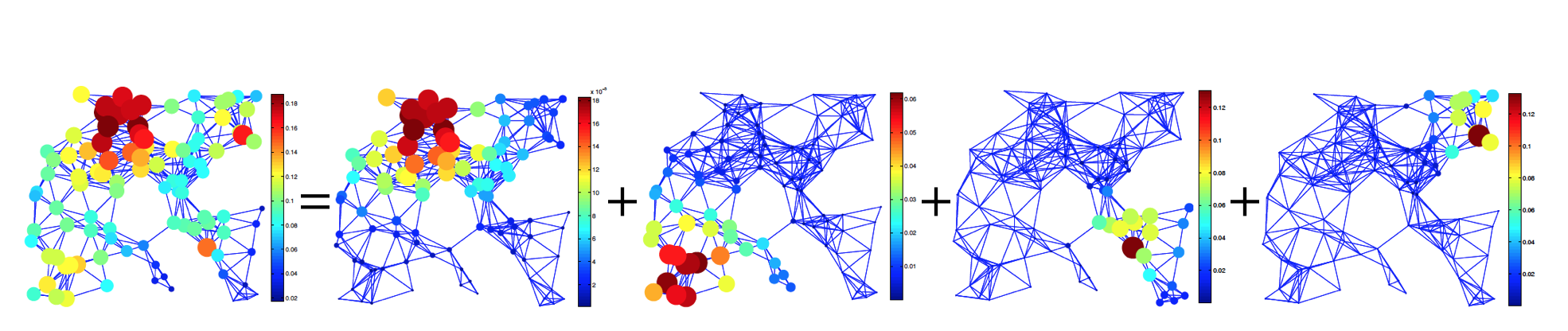}
\label{etex_sparsity}}
	\vspace{-0.2cm}
        \caption{Decomposition of a graph signal (a) in four localized simple components (b), (c), (d), (e). Each component is a heat diffusion process $(e^{-\tau\L})$ at time $\tau$ that has started from different network nodes ($\delta_n$). The size and the color of each ball indicate the value of the signal at each vertex of the graph.}
        \label{fig:diffusion_example}
\end{figure*}

We  then cast a new graph learning problem that aims at estimating a graph that best explains the data measurements, denoted as graph signals,  under the heat diffusion model. Specifically, we represent our signals as a linear combination of a few (sparse) components from a graph dictionary consisting of heat diffusion kernels. The graph learning problem is then formulated as a regularized inverse problem where both the graph and the sparse coefficients are unknown.   We propose a new algorithm to solve the resulting nonconvex optimization problem, which, under mild assumptions~\cite{Bolte_2014}, guarantees that the iterates converge to a critical point. We finally provide a few illustrative experiments on synthetic data, as well as on two  real world datasets that capture (i) the diffusion of tracers in atmospheric systems and (ii) the mobility patterns of Uber trips in New York City.   The graph recovered from the first dataset correctly captures the trajectory of the chemical tracer, while the graphs learned from the Uber data reveal meaningful mobility patterns at different time intervals of the day across the city.  The results confirm that the proposed algorithm is effective at inferring meaningful graph
topologies in both synthetic and real world settings. Our framework is one of the first attempts to learn graphs carrying the structure of data that is not necessarily smooth but instead obeys a more generic sparse model. We believe that this framework will prove particularly useful in the analysis of social and biomedical networks, for example, where the data structure is not immediate from the application settings or design constraints.

The structure of the paper is as follows. We first highlight some related work on the learning of graph topologies in Section \ref{related_work}.   In Section \ref{preliminary_definitions},  we introduce our signal model and the structure of the diffusion dictionary. The graph learning algorithm is presented  in Section \ref{Se:Graph_Learn}.  Finally, in Section \ref{Se:Exp_results}, we evaluate the performance of our algorithm for both synthetic and real world graph signals.

\section{Related work}
\label{related_work}

A number of approaches have recently been proposed to learn the geometry of data. Intense research efforts have been dedicated to methods for estimating covariance matrices (see, e.g.~\cite{Friedman08}), which carry information about the data geometry. Richer structures can be estimated by learning data graphs instead of the mere covariance matrix. For example, the work in \cite{Lake10} learns a valid graph topology (the adjacency matrix) with an optimization problem that is very similar to sparse inverse covariance estimation, but it instead involves a regularized full-rank Laplacian matrix. Then, the authors in~\cite{Dong_2015} relax the assumption of a full-rank matrix and propose to learn a valid graph Laplacian by imposing smoothness of observations on the graph. Thus, instead of focusing on pairwise-correlations between random variables, they explore the link between the signal model and the graph topology to learn a graph that provides a globally smooth representation of the corresponding graph signals. This framework has been extended further to yield a more scalable algorithm for learning a valid graph topology \cite{Kalofolias16}.  The authors in \cite{Pavez16} propose an algorithm to estimate a generalized Laplacian matrix instead of the classical combinatorial or normalized Laplacian. Finally, manifold learning certainly represents another important class of works that aims at estimating the data structure and bears some similarity with the graph learning algorithms in some specific settings \cite{Belkin05}. However, all the above works assume that the data evolve smoothly on the underlying structure, which is not necessarily the ideal model for all datasets. 

The idea of recovering graph topologies for different graph signal models is relatively new and has not yet received a lot of attention. An autoregressive model that is based on graph filter dynamics is  used in \cite{Mei15}  to discover unknown relations among the vertices of a set of  time series. The authors in \cite{PasdeloupGMPR16} model the observations as being measured after a few steps of diffusing signals that are initially mutually independent and have independent entries. The diffusion process is modeled by powers of the normalized Laplacian matrix.  They propose an algorithm for characterizing and then computing a set of admissible diffusion matrices, which relies on a good estimation of the covariance matrix from the independent signal observations. The problem of estimating a topology from signal observations that lead to particular  graph shift operators is studied in  \cite{SegarraMMR16}. The authors propose to learn a sparse graph matrix that can explain signals from graph diffusion processes, under the assumption that eigenvectors of the shift operators, i.e., the graph templates, are already known. The graph learning problem  then becomes equivalent to learning the eigenvalues of the shift matrix. We will discuss the differences with this scheme in the experimental section. Contrary to the existing works, we learn a graph diffusion process without making any assumption on the eigenvectors of the graph process but instead make an explicit assumption on the diffusion process and the sparse signal model. 

\section{Sparse representation of graph signals}
\label{preliminary_definitions}
\subsection{Signal representation}
We consider a weighted and undirected graph $\mathcal{G}=(\V,\E,W)$, where $\V$ and $\E$ represent the vertex (node) and edge sets of the graph, respectively. The $N\times N$ matrix $W$ contains the edge weights, with $W_{ij} =W_{ji} $ denoting the positive weight of an edge connecting vertices $i$ and $j$, and $W_{ij}=0$ if there is no edge. Without loss of generality, we assume that the graph is connected.  The  graph Laplacian operator  is defined as $L=D-W$, where $D$ is the diagonal degree matrix with the $i^{th}$ diagonal element equal to the sum of the weights of all edges incident to vertex $i$ \cite{Chung97}. 
Being a real symmetric and positive semidefinite matrix, the graph Laplacian has an orthonormal basis of eigenvectors. We let $\chi = [\chi_0,\chi_1,...,\chi_{N-1}]$ denote the eigenvector matrix of $L$. The diagonal matrix $\Lambda$ contains the corresponding eigenvalues $0=\lambda_0<\lambda_1\le\lambda_2\le...\le\lambda_{N-1}$ on its diagonal.

A graph signal is a function $x: \V \to \Rbb$
such that $x(v)$ is the 
value of the function at the vertex $v \in \V$.  We consider the factor analysis model from~\cite{Bartholomew11}  
as our graph signal model, which is a generic linear statistical model that aims at explaining observations of a given dimension with a potentially smaller number of unobserved latent variables. Specifically, we consider
\begin{equation}
x = \D h + u_x + \epsilon,
\label{eq:signal_model}
\end{equation}
where $x \in \mathbb{R}^N$ is the observed graph signal, $h \in \mathbb{R}^K$ is the latent variable that controls $x$, and $\D \in \mathbb{R}^{N \times K}$ is a representation matrix that linearly relates the two  variables, with $K\ge N$. The parameter $u_x \in \mathbb{R}^N$ is the mean of $x$, which we set to zero for simplicity, and $\epsilon$ is a multivariate Gaussian noise with zero mean and covariance $\sigma_\epsilon^2I_N$. 

To represent signals residing on graphs, and especially to identify and exploit structure in the data, we need to take the intrinsic geometric structure of the underlying graph into account. This structure is usually incorporated in the columns of the representation matrix, i.e., atoms of a dictionary \cite{Zhang12,Thanou14}. These atoms carry spectral and spatial characteristics of the graph.  Specifically, one can consider  spectral graph dictionaries defined by filtering the eigenvalues of the graph Laplacian in the following way:  
\begin{equation}
\D= [\widehat{g}_1(L) ~\widehat{g}_2(L)\dots \widehat{g}_S(L)   ],
 \label{eq:generating_process}
 \end{equation}
where $\{\widehat{g}_s(\cdot)\}_{s=1,\ldots,S}$  are graph filter functions defined on a domain containing the spectrum of the graph Laplacian. Each of these filters captures different spectral characteristics of the graph signals. 

For efficient signal representation, the latent variables $h$ should be sparse such that they reveal  the core components
of the graph signals \cite{Rubinstein2010}. In particular, one can impose  a Laplace (sparse) prior on the latent variable $h$ like
\begin{equation}
p(h) = \prod_i \alpha~\text{exp}(-\alpha|h(i)|),
\label{eq:ph_graph}
\end{equation}
where $\alpha$ is constant,
and a Gaussian prior on the noise $\epsilon$. Then the conditional probability of $x$ given $h$ can be written as
\begin{equation*}
p(x|h) \sim \mathcal{N}(\D h, \sigma_\epsilon^2I_N).
\label{eq:pxcon}
\end{equation*}
Given the observation $x$ and the Laplace  prior distribution of $h$ in Eq.~(\ref{eq:ph_graph}), we can compute  a maximum a posteriori (MAP) estimate of the sparse set of components. Specifically, by applying Bayes' rule and assuming without loss of generality that $u_x = 0$, the MAP estimate of the latent variable $h$ is \cite{Gribonval11}:
\begin{equation}
\begin{split}
h_\text{MAP}(x) \coloneqq &\arg \max_h p(h|x) 
= \arg \max_h p(x|h)p(h) \\
= &\arg \min_h \left(-\text{log}~p_E(x-\D h) -\text{log}~p_H(h)\right)\\
= & \arg \min_{h} \|x - \D h\|_2^2 + \alpha \|h\|_1,
\end{split}
\label{eq:map}
\end{equation}
where $\|\cdot\|_1$ denotes the $\ell^1$-norm. 

 Sparsity-based inverse problems have been widely used in the literature to perform classical signal processing tasks on the observations $x$, such as denoising and inpainting.  
Sparsity however largely depends on the
design of the dictionary, which itself depends on the graph.
In the following, we discuss the choice of the representation matrix and the latent variables in our heat diffusion signal model.  

\subsection{Diffusion signals on graphs}
\label{diffusion_of_signals}
In this paper, we focus on graph signals generated from heat diffusion processes, which are useful in identifying processes evolving nearby a starting seed node.   In particular, the graph Laplacian matrix is used to model the diffusion of the heat throughout a graph or, more generally, a geometric manifold. The flow of the diffusion is governed by the following second order differential equation with initial conditions: 
\begin{equation}
\label{differential_equation}\frac{\partial x}{\partial \tau} - L x = 0, ~ x(v, 0) = x_0(v)
\end{equation}
where $x(v, \tau)$ describes the heat at node $v$ at time $\tau$, beginning from an initial distribution of heat given by $x_0(v)$ at time zero. The solution of the differential equation is given by 
\begin{equation}
x(v, \tau) = e^{-\tau L} x_0(v).
\end{equation}

Going back to our graph signal model, the graph heat diffusion operator is defined as \cite{Smola03}
\begin{equation*}
\widehat{g}(L):= e^{-\tau L}=\chi e^{-\tau\Lambda}\chi^T.
 \label{heat_diffusion}
 \end{equation*}
Different powers $\tau$ of the heat diffusion operator 
correspond to different rates of heat flow over the graph. If such operators are used to define a dictionary in (\ref{eq:generating_process}), our graph signal model of (\ref{eq:signal_model}) becomes
\begin{equation*}
x=\D h + \epsilon= [e^{-\tau_1 L} ~ e^{-\tau_2 L} ~ \cdots ~ e^{-\tau_S L} ~ ]h + \epsilon,
\end{equation*} 
which is a linear combination of different heat diffusion processes evolving on the graph. For each diffusion operator $e^{-\tau_s L}$, the signal component $e^{-\tau_s L} h_s$  can also be interpreted as the result of filtering an initial graph signal $h_s$ with an exponential, low-pass filter $e^{-\tau_s L}$ on the graph spectral domain. The obtained signal $x$ is the sum of each of these simple components $x = \sum_{s=1}^S e^{-\tau_s L}h_s$. Notice that the parameter $\tau$ in our model carries a notion of scale. In particular, when $\tau $ is small, the $i^{th}$ column of $\D$, i.e., the atom $\D(i,:)$ centered at node $i$ of the graph is mainly localized in a small neighborhood of $i$. As $\tau$ becomes larger, $\D(i,:)$ reflects information about the graph at a larger scale around $i$. Thus, our signal model can be seen as an additive model of diffusion processes that started at different time instances. Finally, the sparsity assumption of Eq. (\ref{eq:ph_graph}) on the latent variables $h$ implies that we expect the diffusion process to start from only a few nodes of the graph and spread over the entire graph over time.  

\section{Learning graph topologies under sparse signal prior}
\label{Se:Graph_Learn}
In many applications, the graph is not necessarily known, and thus the MAP estimate of the latent variables in (\ref{eq:map}) cannot be solved directly.  In the following, we show how the sparse representation model of the previous section can be exploited to infer the underlying graph topology, under the assumption that the signals are generated by a set of  heat diffusion processes. 
First, we formulate the graph learning problem, and then we propose an efficient algorithm to solve it.  

\subsection{Problem formulation}
Given a set of $M$ signal observations $X=[x_{1},x_{2},...,x_{M}]\in \Rbb^{N\times M}$,  resulting from heat diffusion processes  evolving on an unknown weighted graph $\mathcal{G}$, our objective is twofold: (i) infer the graph of $N$ nodes by learning the graph Laplacian $L$, and  (ii) learn,  for each signal, the latent variable  that reveals the sources of the observed processes, i.e., $H = [h_1, h_2, ...,h_M]$ and the diffusion parameters $\tau = [\tau_1, \tau_2, ...,\tau_S]$.   As the graph Laplacian $L$ captures the sparsity pattern of the graph, learning $L$ is equivalent\footnote{Since our graph does not contain self-loops, the weight matrix $W$ of the graph can be simply computed as $W = -L$, and then setting the diagonal entries to zero.} to learning the graph $\mathcal{G}$.  This results in the following joint optimization problem for $H$, $L$, and $\tau$:
\begin{align}
\label{eq:opt_prob_heat_kernel}
 \underset{L ,~H, ~\tau}{\mbox{minimize}} ~~~ & \| X -\D H \|^{2}_{F} +  \alpha \sum_{m = 1}^M\|h_m\|_1 + \beta \|L\|_F^2 \\ 
 \mbox{subject to}  
  &  ~~  \D = [e^{-\tau_1 L} ~ e^{-\tau_2 L} \dots e^{-\tau_S L} ~ ]\nonumber \\
\nonumber ~~~ & \mbox{tr}(L)=N, \\
\nonumber ~~~ & L_{ij} = L_{ji} \le 0, ~i \neq j, \\
\nonumber ~~~ & L \cdot \mathbf{1}= \mathbf{0},\\
\nonumber ~~~ & \tau \ge 0,
\end{align} 
where $h_m$ corresponds to the $m^{th}$ column of the matrix $H$. According to Eq.~(\ref{eq:map}), the objective can be interpreted as the negative log-likelihood of the latent variables (columns of $H$) conditioned on the graph Laplacian $L$. 
The positive scalars $\alpha$ and $\beta$ are regularization parameters, while $\mathbf{1}$ and $\mathbf{0}$ denote the vectors of all ones and zeros, respectively. In addition, $\mathrm{tr}(\cdot)$ and $\|\cdot\|_F$ denote the trace and Frobenius norm of a matrix, respectively. The trace constraint  acts as a normalization factor that fixes the volume of the graph and the remaining constraints guarantee that the learned $L$ is a valid Laplacian matrix that is positive semidefinite.
Note that the trace constraint, together with the other constraints, also fixes the $\ell^1$-norm of $L$, while the Frobenius norm is added as a penalty term in the objective function to control the distribution of the off-diagonal entries in $L$, that is, the edge weights of the learned graph. When $L$ is fixed, the optimization problem bears similarity to the linear combination of $\ell^1$ and $\ell^2$ penalties in an elastic net regularization~\cite{Zou05}, in the sense that the sparsity term is imposed by the trace constraint. When $L$, $\tau$ are fixed, problem (\ref{eq:opt_prob_heat_kernel}) becomes equivalent to a MAP estimator, as discussed in the previous subsection. 

Note that our problem formulation depends on the number of blocks $S$, i.e., the number of scales of the diffusion processes. The choice of $S$ depends on the training signals, in particular, on the number of scales that one can detect in the training data. As we expect the diffusion processes to be localized, we typically choose a small value for $S$, say, 1 to 3. One of the blocks would correspond to a very small scale (i.e., highly localized atoms), and the other blocks would capture larger scale, but still somewhat localized patterns. 

The optimization problem~\eqref{eq:opt_prob_heat_kernel} is nonconvex with respect to $L, H, \tau$ simultaneously. In particular, the data fidelity term $\| X -\D H \|^{2}_{F}$ is smooth but nonconvex as it contains the product of the three matrix variables (e.g., $e^{-\tau L}H$).  As such, the problem may have many local minima and  solving it is hard. One could apply alternating minimization, where at each step of the alternation we update one variable by fixing the rest. This, however, does not provide convergence guarantees to a local minimum and, moreover, solving the problem with respect to $L$ is difficult due to the matrix exponential, which makes the problem nonconvex even when $\tau, H$ are fixed. In the next section, we propose an effective algorithm to solve the graph learning problem, which is not affected by this difficulty.

\subsection{Graph learning algorithm}
\label{sec:graph_learning}
In order to solve \eqref{eq:opt_prob_heat_kernel}, we apply a proximal alternating linearized minimization algorithm (PALM)~\cite{Bolte_2014}, which can be interpreted as alternating the steps of a proximal forward-backward scheme \cite{Bauschke2011}.
PALM is a general algorithm for solving a broad class of nonconvex and nonsmooth minimization problems, which, under mild assumptions~\cite{Bolte_2014}, guarantees that the iterates converge to a critical point. Moreover, it does not require convexity of the optimization problem with respect to each variable separately.  The basis of the algorithm is alternating minimization between the three variables $(L, H, \tau)$, but in each step  we linearize the nonconvex fitting term $\|X -\D H\|_{F}^2$ with a first order function at the solution obtained from the previous iteration. In turn, each step becomes the proximal regularization of the nonconvex function, which can be solved efficiently.   More specifically, the algorithm consists of three main steps: (i) update of $H$, (ii) update of $L$, (iii) update of $\tau$, and inside each of these steps we compute the gradient and estimate the Lipschitz constant with respect to each of the variables.  Algorithm~\ref{Alg_Heat_Kernel} contains a summary of the basic steps of PALM adapted to our graph learning problem.

\begin{algorithm}[t]
\caption{Learning heat kernel graphs (\textbf{LearnHeat})}
\begin{algorithmic}
\item [ 1:] {\bf Input:} Signal set $X$,  number of iterations $\mathsf{iter}$
\item [ 2:] {\bf Output:}  Sparse signal representations $H$, graph Laplacian $L$,  diffusion parameter $\tau$
\item [ 3:] {\bf Initialization:} $L=L^{0}$, $\D^0= [e^{-\tau_1 L} ~ e^{-\tau_2 L} \dots e^{-\tau_S L} ~ ]$
\item [ 4:] {{\bfseries for} $t=1,2,...,\mathsf{iter}$ \bfseries{do}:}
\item [ 5:]  \quad {Choose $c_t = \gamma_1 C_1(L^{t}, \tau^t)$} 
\item [ 6:]  \quad  \quad Update $H^{t+1}$ by solving opt. problem (\refeq{eq:updateH})
\item [ 7:]  \quad {Choose $d_t = \gamma_2 C_2(H^{t+1}, \tau^t)$} 
\item [ 8:]  \quad  \quad (a) Update $L^{t+1}$ by solving opt. problem (\refeq{eq:updateLQP})
\item [ 9:]  \quad \quad (b) Update $\D^{t+1}=[e^{-\tau_1^{t} L^{t+1}}\dots e^{-\tau_S^{t} L^{t+1}}]$
\item [10:]  \quad {Choose $e_t = \gamma_3 C_3(L^{t+1}, H^{t+1})$} 
\item [11:]  \quad \quad (a) Update $\tau^{t+1}$ by solving opt. problem (\ref{eq:update_tau})
\item [12:]  \quad \quad (b) Update  $\D^{t+1}=[e^{-\tau_1^{t+1} L^{t+1}}\dots e^{-\tau_S^{t+1} L^{t+1}}]$
\item [13:] {\bf end for}  
\item [14:]  $L=L^{\mathsf{iter}}, H=H^{\mathsf{iter}}, \tau = \tau^{\mathsf{iter}}$.
\end{algorithmic}
\label{Alg_Heat_Kernel}
\end{algorithm}

In the following, we  explain in detail each of the steps of Algorithm~\ref{Alg_Heat_Kernel}. We make use of the following definitions: 
 $$Z(L, H, \tau) = \|X -\D H\|_{F}^2, \quad
 f(H) = \sum_{m=1}^M\|h_m\|_1,$$ $$ g(L) = \delta(L| \mathcal{C})+\beta \|L\|_F^2,$$ 
 where $\delta$ is an indicator function for the convex set $\mathcal{C} = \{\mbox{tr}(L) = N,  L_{ij} = L_{ji} \le 0, i \neq j, L \cdot \mathbf{1}= \mathbf{0}\}$, defined as
\begin{equation*}
\delta(L| \mathcal{C})=
    \begin{cases}
            1, &         \text{if } L\in\mathcal{C}\\
            +\infty, &         \text{ otherwise}.
    \end{cases} 
  \end{equation*}

\subsubsection{Update of $H$ (Algorithm \ref{Alg_Heat_Kernel}: lines 5-6)}
For iteration $t+1$ of the sparse coding update step, we solve the following optimization problem:
\begin{align}
\nonumber H^{t+1} = \underset{ H}{\operatorname{argmin~}} &\langle H-H^{t}, \nabla Z_{H}(L^{t}, H^t, \tau^t) \rangle \\
&+ \frac{c_t}{2}\|H - H^t\|^2 + f(H),
\label{eq:updateH}
\end{align}
where $L^t, H^t, \tau^t$ are the updates obtained at iteration $t$ and $c_t$ is a positive constant. This step can simply be viewed as the proximal regularization of the nonconvex function $Z(L, H, \tau)$, linearized at $(L^t, H^t, \tau^t)$:  
\begin{equation*}
 H^{t+1} = \prox_{c_t}^f \big( H^t - \frac{1}{c_t}\nabla_{H}Z(L^{t}, H^{t}, \tau^{t}) \big)
\end{equation*}
where $\prox_{c_t}^f$ is the proximal operator of the convex function $f(H)$ with parameter $c_t$, given by  
\begin{equation}
 \prox_{c_t}^f(z) = 
            \mbox{sign}(z) \mbox{max}(|z| - \alpha/c_t, 0), 
\end{equation}
where $z = H^t - \frac{1}{c_t}\nabla_{H}Z(L^{t}, H^{t}, \tau^{t})$. The required gradient of $Z(L, H, \tau) = \|X -\D  H\|_{F}^2$ with respect to $H$ is computed in Appendix~\ref{appendixA_H}. The parameter $c_t$ is defined such that $c_t= \gamma_1 C_1(L^{t}, \tau^t)$, with $\gamma_1>1$ and the Lipschitz constant $C_1(L^{t}, \tau^t)$ of  $\nabla_{H} Z(L^t, H, \tau^t)$ with respect to $H$, as derived  in Appendix \ref{appendixB_H}.

\subsubsection{Update of $L$ (Algorithm \ref{Alg_Heat_Kernel}: lines 7-9)}
The graph update step is performed by 
\begin{equation}
 L^{t+1} = \prox_{d_t}^g \big( L^t - \frac{1}{d_t}\nabla_{L}Z( L^{t},H^{t+1}, \tau^{t}) \big), 
 \label{eq:updateL}
\end{equation}
with $d_t= \gamma_2 C_2(H^{t+1}, \tau^t)$ for some $\gamma_2>1$ and the estimate $C_2(H^{t+1}, \tau^t)$ of the Lipschitz constant of $\nabla_L Z(L, H^{t+1}, \tau^t)$ described in Appendix \ref{appendixB_L}.
Given that $g(L) = \delta(L| \mathcal{C})+\beta \|L\|_F^2$ comprises a quadratic term constrained in a convex polytope, the proximal minimization step \eqref{eq:updateL} is a quadratic program (QP) that can be written as:
\begin{align}
\label{eq:updateLQP}
\nonumber \underset{L }{\mbox{minimize}} ~~~ & \langle L - L^t, \nabla_L Z(L^{t}, H^{t+1}, \tau^{t}) \rangle  \\ \nonumber &  ~~ + \frac{d_t}{2}\| L - L^{t} \|^{2}_{F} + \beta \|L\|_F^2 \\ 
 \mbox{subject to}  
\nonumber ~~~ & \mbox{tr}(L)=N, \\
~~~ & L_{ij} = L_{ji} \le 0, ~i \neq j, \\
~~~ & L \cdot \mathbf{1}= \mathbf{0}.
\nonumber
\end{align} 
This requires the gradient of $Z(L, H, \tau) = \|X -\D H\|_{F}^2$ with respect to $L$, the derivation of which can be found in Appendix~\ref{appendixA_L}. Given  this gradient, the optimization problem~\eqref{eq:updateLQP} can be solved using operator splitting methods, such as the alternating direction method of multipliers (ADMM) \cite{Boyd11}. In this paper, we solve the problem by using the algorithm proposed in \cite{ODonoghue}, which converts the problem to a convex cone optimization problem, and utilizes ADMM to solve the homogenous self-dual embedding. Compared to other methods, this approach finds both primal and dual solutions of the problem, is free of parameters, and scales to large problem sizes. 

\subsubsection{Update of $\tau$ (Algorithm \ref{Alg_Heat_Kernel}: lines 10-12)}

Finally, we can update the diffusion parameters $\tau = [\tau_1, \tau_2, ..., \tau_S]$ following the same reasoning as above. The corresponding optimization problem can be written as
\begin{align}
\nonumber \underset{\tau }{\mbox{minimize}} ~~~ & \langle \tau - \tau^t, \nabla_\tau Z(L^{t+1}, H^{t+1}, \tau^t) \rangle + \frac{e_t}{2}\| \tau - \tau^{t} \|^{2}_{F}  \\ 
 \mbox{subject to}  
\label{eq:update_tau} ~~~ &\tau \ge 0,
\end{align} 
where $e_t = \gamma_3 C_3(H^{t+1}, L^{t+1})$, with $\gamma_3>1$ and the Lipschitz constant $C_3(H^{t+1}, L^{t+1})$  computed in Appendix \ref{appendixB}. This problem has a closed form solution given by 
\begin{equation}
\tau^{t+1} = 
             \mbox{max}\big(- \frac{ \nabla_\tau Z(L^{t+1}, H^{t+1}, \tau^t) - e_t \tau^t}{e_t}, 0\big),
\end{equation}
with the gradient computed in Appendix~\ref{appendixA_tau}.
Finally, we note that if we have an a priori estimate of the diffusion parameters $\tau$ (e.g., from the training phase) then we solve our optimization problem with respect to $L, H$ by following the first two steps of our algorithm. 

\subsection{Discussion on the computational complexity}
In the following, we discuss the computational complexity our graph learning algorithm. 
%
Dealing with the heat diffusion processes $e^{-\tau_s L}$ represents one of the main computational bottlenecks. Both, the computation of the matrix exponential via a spectral decomposition or via the scaling and squaring method~\cite{Higham:2008} as well as the computation of its gradient described in Appendix~\ref{appendixA_L} require $\mathcal{O}(N^3)$ operations. Thus, this part of the algorithm can be expected to become time consuming for very large graphs. One way to reduce this cost is to approximate the heat diffusion kernel with a polynomial of degree $K$, reducing the complexity of applying a heat diffusion process to $\mathcal{O}(|\E| K)$, where $|\E|$ is the number of edges of the graph Laplacian. Since we generally consider heat diffusion processes that remain well localized, the degree $K$ will typically be small. This approximation of the heat diffusion process is particularly efficient when the graph Laplacian is sparse. Also, it can be expected that the complexity of the gradient computation greatly reduces when using a polynomial approximation of the kernel; see~\cite{Deadman16} for some recent work in this direction. A detailed investigation of this aspect is part of our future work. 

We further note that the computational complexity of the sparse coding step (lines 5-6 of the Algorithm) is dominated by the cost of computing the Lipschitz constant (see Appendix \ref{appendixA_H}), which requires the computation of the product $\D^T\D$ and is of order $\mathcal{O}(S^2N^3)$. Again, this cost greatly reduces when using a polynomial approximation of the kernel. The update of the sparse codes in  Eq. (\refeq{eq:updateH}) requires $\mathcal{O}(N^2S)$ operations. Finally, the  update of the graph Laplacian (Algorithm \ref{Alg_Heat_Kernel}: lines 7-9) consists of three steps: the recursive approximation of the Lipschitz constant (Appendix \ref{appendixB_L}), the computation of the gradient discussed above and the solution of the optimization problem~(\refeq{eq:updateLQP}).
The solution of~(\refeq{eq:updateLQP}) involves three main steps  \cite{ODonoghue}, among which the most expensive one is solving a linear system. For large scale systems, this can be done efficiently by applying a conjugate gradient method.  Finally, the computation of the Lipschitz constant in the update of $\tau$ (see Appendix \ref{appendixB_tau}) requires the computation of the spectral norm of $L$, which can be estimated in $\mathcal{O}(|\E|)$ operations by a few steps of the power or Lanczos method~\cite{Golub2013}.

\section{Experiments}
\label{Se:Exp_results}

In this section, we evaluate the performance of the proposed algorithm in both synthetic and real world experiments. We solve the optimization problem of Eq. (\refeq{eq:updateLQP}) using ADMM, which is implemented with the splitting conic solver \cite{ODonoghue}, a numerical optimization package for solving large-scale convex cone problems\footnote{The conic solver can be found at https://github.com/cvxgrp/scs}.    As a termination criteria, we stop the algorithm when a maximum number of iterations (set to 1000 in our experiments) is reached or the absolute difference in the value of the objective function at two consecutive iterations is smaller than $10^{-4}$. 

\subsection{Results on synthetic data}  
\subsubsection{Simulation settings}
We first test the performance of the learning algorithm by comparing the learned graph to the one from the groundtruth in synthetic datasets. 
We evaluate the performance of the algorithm on random graphs of $N=20$ vertices, generated from three different models: the RBF random graph,  the Barab\'{a}si-Albert model (BA) \cite{Barabasi99}, and the  Erd\H{o}s-R\'{e}nyi model (ER) \cite{Erdos60}.  In the case of the RBF graph, we generate the coordinates of the vertices uniformly at random in the unit square, and we set the edge weights based on a thresholded Gaussian kernel function so that $ W(i,j)=e^{-\frac{[\text{dist}(i,j)]^2}{2\sigma^2}}$ if the distance between vertices $i$ and $j$ is less than or equal to $\kappa$, and zero otherwise. We further set $\sigma = 0.5$ and $\kappa = 0.75$ in our experiments. In the ER graph, an edge is included with probability 0.2 independently of the other edges. Finally, in the BA graph, we add vertices one after the others and connect them to existing vertices following a preferential attachment mechanism. Given the adjacency matrix of each type of graph, we finally compute the graph Laplacian and we normalize in such a way that its trace is equal to $N$.

%

With the above model-based graphs, we then construct synthetic graph signals as follows. We use the graph Laplacian to generate an oracle dictionary of the form $\D= [e^{-\tau_1 L} ~e^{-\tau_2 L} ]$, with $\tau_1 =2.5, \tau_2 = 4$, for the RBF and the ER graph and  $\tau_1 = 1, \tau_2 = 4$ for the BA model. These values are chosen in such a way that our dictionaries contain two patterns that are sufficiently distinct from each other. In particular, the one corresponding to a small $\tau$ captures a very localized pattern while the one corresponding to a large $\tau$ captures a diffusion that has already spread in the local neighborhood of the vertex. 
We then generate 100 graph signals by linearly combining three random atoms from the dictionary with random coefficients drawn from a Gaussian distribution with zero mean and unit variance. The sparse vector of random coefficients represents the initial heat on the graph. We finally hide the graph structure, and apply Algorithm \ref{Alg_Heat_Kernel} with different sets of parameters $\alpha = [10:10^{-0.5}:10^{-6}]$ and $\beta =  [1:10^{-1}:10^{-2}]$ in order to estimate the graph only from the signal observations. The initialization of the graph Laplacian is done with a random valid Laplacian matrix. We compare our algorithm (LearnHeat) with the following two methods: (i) the algorithm proposed in \cite{SegarraMMR16}, which is based on learning the graph Laplacian from diffusion filters and does not have any assumption on the global smoothness of the signals, and (ii) the algorithm in \cite{Dong_2015} which recovers a graph Laplacian by assuming  global smoothness of the signals on the graph. We solve the algorithm in \cite{SegarraMMR16} for different values of the parameter $\epsilon =  [0:0.02:2]$, where $\epsilon$  controls the imperfection of the spectral templates estimated from the covariance matrix, and provides a constraint on how close the optimized matrix should be to these templates. We threshold the learned Laplacian matrices by ignoring entries whose absolute values are smaller than $10^{-4}$. 

\begin{figure*}[]
      \centering    
       \subfigure[Gaussian RBF: Groundtruth]
{ \includegraphics[width=4cm]{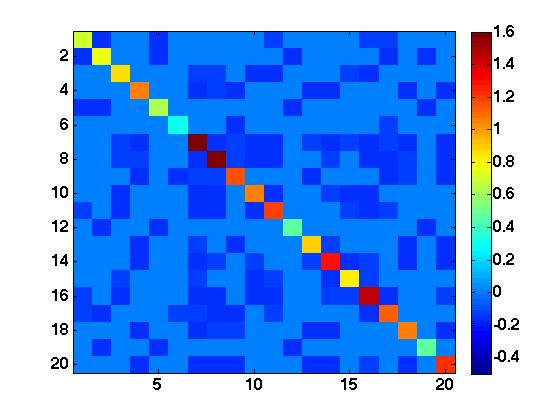}   
\label{laplacian_gt_rbf}}
\subfigure[\textbf{LearnHeat}]
{ \includegraphics[width=4cm]{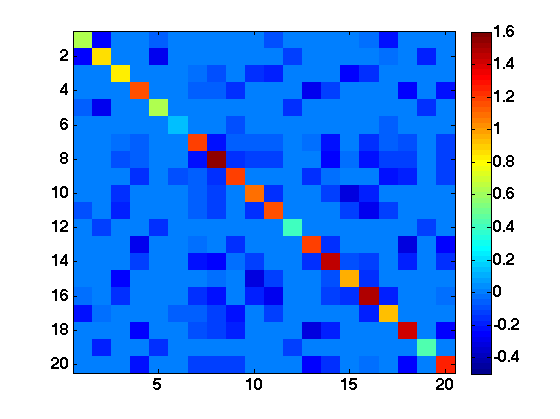}
\label{laplacian_gaussian_rbf}}
\subfigure[\textbf{Diffusion filters \cite{SegarraMMR16}}]
{ \includegraphics[width=4cm]{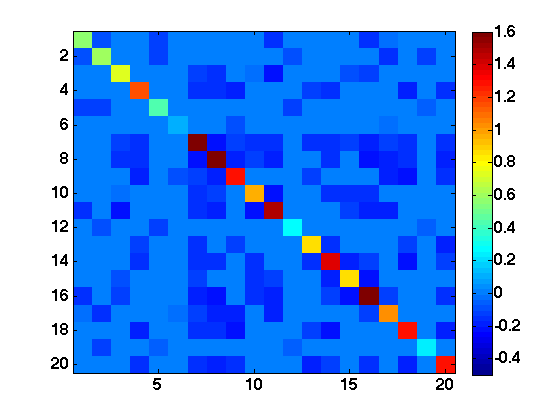}
\label{Sant_gaussian_rbf}}
\subfigure[\textbf{Smooth model \cite{Dong_2015}}]
{ \includegraphics[width=4cm]{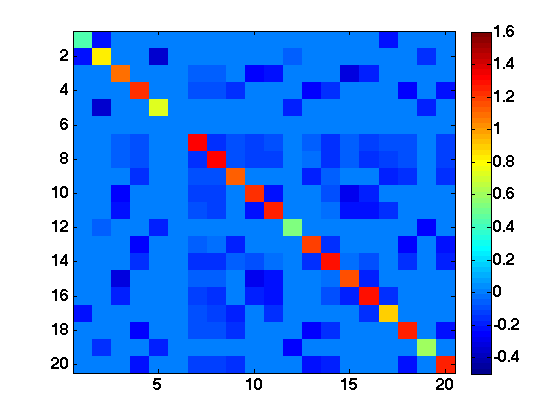}
\label{Smooth_gaussian_rbf}}

            \subfigure[Gaussian RBF: Groundtruth]
{ \includegraphics[width=4cm]{gaussian_groundtruth.png}   
\label{laplacian_gt_rbf}}
\subfigure[\textbf{LearnHeat}] 
{ \includegraphics[width=4cm]{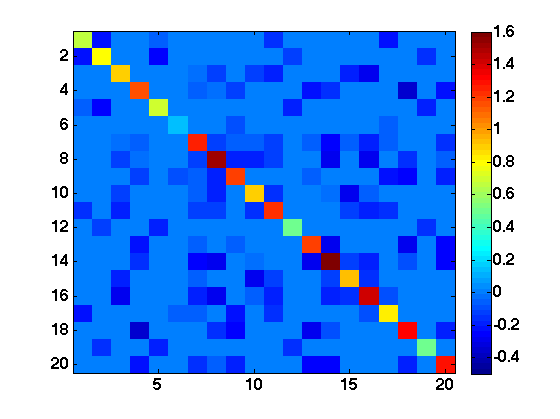}
\label{laplacian_gaussian_rbf}}
\subfigure[\textbf{Diffusion filters \cite{SegarraMMR16}}] 
{ \includegraphics[width=4cm]{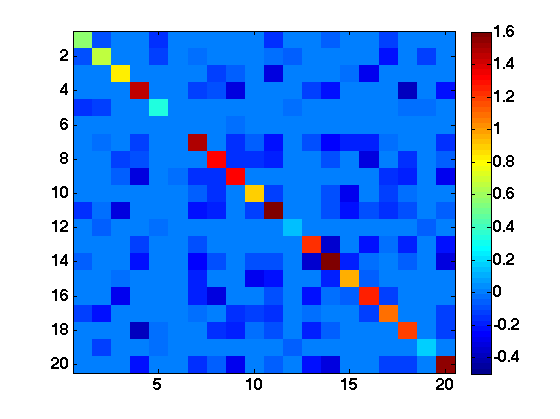}
\label{Sant_gaussian_rbf}}
\subfigure[\textbf{Smooth model \cite{Dong_2015}}]
{ \includegraphics[width=4cm]{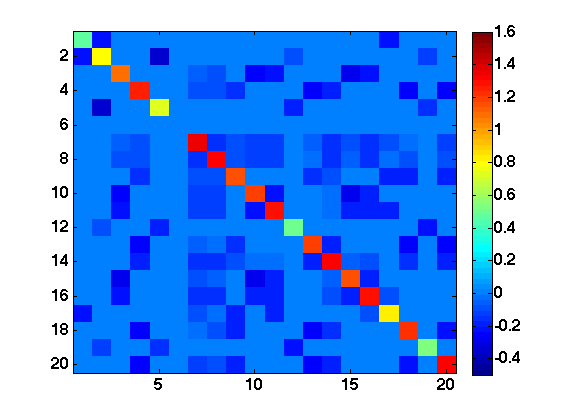}
\label{Smooth_gaussian_rbf}}
	\vspace{-0.2cm}
        \caption{The learned graph Laplacian matrices for a Gaussian RBF graph. The color indicates the values of the entries of the graph Laplacians.  The first row illustrates the groundtruth Laplacian and the Laplacians recovered with LearnHeat, the algorithm of \cite{SegarraMMR16}  and the smooth signal model of \cite{Dong_2015}, when the training signals are clean. The second row illustrates the same results obtained from noisy training signals.}
        \label{fig:syn_visual_heat_kernel}
\end{figure*}

\subsubsection{Graph learning performance}
We first compare visually the learned graph Laplacian for an RBF graph model with the corresponding groundtruth one.   The results illustrated in Fig. \ref{fig:syn_visual_heat_kernel} are the ones obtained for the pair of $\alpha$ and $\beta$ that leads to the best quantitative results (see below), and the best $\epsilon$ for \cite{SegarraMMR16}. First, we consider the noiseless case of clean training signals (first row). We observe that both the graph Laplacians learned with the proposed algorithm and the diffusion filters of \cite{SegarraMMR16} are visually consistent with the groundtruth Laplacian, reaching an \textit{F-measure} score of  0.9784 and 0.9927 respectively. On the other hand, the performance of \cite{Dong_2015} that is based on a smooth signal model is worse in terms of  \textit{F-measure}  score (0.9173). This is quite expected as, when the diffusion parameter $\tau$ is relatively small,  the signals generated by heat diffusion processes consist mainly of localized, piecewise smooth components that can be better captured with the other two algorithms. A globally smooth signal model can help recovering parts of the graph, but is not accurate enough to reveal the true topology.  However, as $\tau$ increases the diffusion tends to a steady state that is a smooth signal on the graph. In that case, the behavior of our algorithm is expected to be close to the one in \cite{Dong_2015}. 

In the second set of experiments, we test the sensitivity of the three algorithms to noise on the training signals. In particular, we add some white noise with zero mean and variance 0.02 to our training signals, leading to a signal to noise ratio of approximately 13 dB. In the second row of Fig. \ref{fig:syn_visual_heat_kernel}, we observe that LearnHeat is quite resilient to the noise, reaching an \textit{F-measure}  score of 0.9552 and an error of 0.2642 in terms of the Frobenius difference of the edge weights compared to the groundtruth ones.  The performance of \cite{SegarraMMR16} seems to deteriorate significantly due to the noise, achieving an  \textit{F-measure}  score of 0.8451 and error weight of 0.3546. This is quite expected as the algorithm is based on the estimation of the eigenvectors of the Laplacian, which  depends on the covariance of the noisy training set. The performance of \cite{Dong_2015}  deteriorates too but less significantly, as this algorithm contains a term in the optimization problem that performs denoising of the training signals. 

In order to evaluate quantitatively the performance of our learning algorithm in recovering the edges of the groundtruth graph, we report the \textit{Precision}, \textit{Recall}, \textit{F-measure} and \textit{Normalized Mutual Information (NMI)} \cite{Manning08} scores, as well as the difference in terms of the Frobenius norm of the edge weights, averaged over ten random instances of  three graph models with their corresponding 100 signal observations. For computing the \textit{NMI}, we first compute a 2-cluster partition of all the vertex pairs using the learned graph, based on whether or not there exists an edge between the two vertices. We then compare this partition with the 2-class partition obtained in the same way using the groundtruth graph.  The LearnHeat results shown in Tables \ref{tab:syn_perf_heat_kernel_noiseless},  \ref{tab:syn_perf_heat_kernel} are the ones corresponding to the best combinations of $\alpha$ and $\beta$ in terms of \textit{F-measure} for noiseless and noisy training signals respectively, while the results of \cite{SegarraMMR16} are the ones obtained for the constant $\epsilon$ that gives the best \textit{F-measure}.   These results confirm that our algorithm is able to learn graph topologies that are very similar to the groundtruth ones and its performance is quite robust to noise. The algorithm of \cite{SegarraMMR16} seems to perform very well in the noiseless case. However, its performance deteriorates in most of the noisy cases. As expected, the worst performance is observed in \cite{Dong_2015}. Since our training signals consist of localized heat diffusion patterns, the performance of \cite{Dong_2015} is significantly penalized  by the global smoothness constraint that is imposed in the optimization problem.

\begin{table*}[]
\small{
\caption{Graph learning performance for clean data}
\centering
\begin{tabular}{c c c c c c c}
\hline\hline
Graph model& \textit{F-measure} & \textit{Precision} & \textit{Recall}  & \textit{NMI} & \textit{$\ell^2$ weight error}  \\ [0.5ex] 
\hline
Gaussian RBF  (LearnHeat)& 0.9779 & 0.9646 & \textbf{0.9920} &  0.8977 & 0.2887\\
Gaussian RBF \cite{SegarraMMR16} & \textbf{0.9911}& \textbf{0.9905} & 0.9919&  \textbf{0.9550}& \textbf{0.2081}\\
Gaussian RBF   \cite{Dong_2015} & 0.8760 & 0.8662 & 0.8966 &  0.5944 & 0.4287\\
\hline
ER  (LearnHeat)& \textbf{0.9303}& \textbf{0.8786} & \textbf{0.9908}& \textbf{0.7886}& \textbf{0.3795}\\
ER  \cite{SegarraMMR16} &0.8799 & 0.8525 & 0.9157&  0.65831 & 0.3968\\
ER   \cite{Dong_2015} & 0.7397 & 0.6987  & 0.8114 &  0.4032 & 0.5284\\
\hline
BA (LearnHeat)& \textbf{0.9147}& \textbf{0.8644} & \textbf{0.9757} &  \textbf{0.7538} & 0.4009\\
BA  \cite{SegarraMMR16} &0.8477& 0.7806 & 0.9351&  0.6009 & \textbf{0.3469}\\
BA   \cite{Dong_2015} & 0.6969 & 0.6043  & 0.8459 &  0.3587& 0.5880\\[1ex]
\hline
\end{tabular}
\label{tab:syn_perf_heat_kernel_noiseless}

}
\end{table*}

\begin{table*}[]
\small{
\caption{Graph learning performance for noisy data}
\centering
\begin{tabular}{c c c c c c c}
\hline\hline
Graph model& \textit{F-measure} & \textit{Precision} & \textit{Recall}  & \textit{NMI} & \textit{$\ell^2$ weight error}  \\ [0.5ex] 
\hline
Gaussian RBF  (LearnHeat)& \textbf{0.9429} & \textbf{0.9518} & \textbf{0.9355} &  \textbf{0.7784} & \textbf{0.3095}\\
Gaussian RBF \cite{SegarraMMR16} & 0.8339 & 0.8184 & 0.8567&  0.5056 & 0.3641\\
Gaussian RBF   \cite{Dong_2015} & 0.8959 & 0.7738  & 0.9284 &  0.5461 & 0.4572\\
\hline
ER  (LearnHeat)& \textbf{0.8217}& 0.7502 & \textbf{0.9183} &  \textbf{0.5413} & \textbf{0.3698}\\
ER  \cite{SegarraMMR16} &0.8195 & \textbf{0.7662} & 0.8905&  0.5331 & 0.3809\\
ER   \cite{Dong_2015} & 0.6984 & 0.5963  & 0.8690 &  0.3426 & 0.5172\\
\hline
BA (LearnHeat)& 0.8155& 0.7503 & 0.8986 &  0.5258 & 0.4036\\
BA  \cite{SegarraMMR16} &\textbf{0.8254} & \textbf{0.7613} & \textbf{0.9068}&  \textbf{0.5451} & \textbf{0.3980}\\
BA   \cite{Dong_2015} & 0.7405 & 0.6800  & 0.8230 &  0.3980& 0.5899\\[1ex]
\hline
\end{tabular}
\label{tab:syn_perf_heat_kernel}

}
\end{table*}

\subsubsection{Algorithm analysis}

\begin{figure}
      \centering
            { \includegraphics[width=7cm]{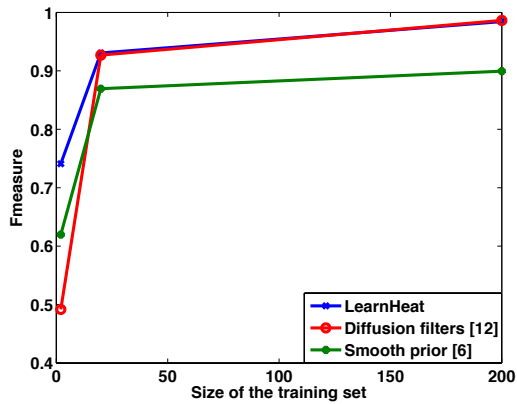}}
	\vspace{-0.2cm}
        \caption{Dependence of the \textit{F-measure}  on the size of the training set for the three different algorithms i.e., LearnHeat, Diffusion Filters and Smooth priors.}
        \label{fig:training_set}
\end{figure}

To understand the effect of the number of the training signals in the learning performance, we run a set of experiments on some clean training signals. In Fig. \ref{fig:training_set}, we illustrate the best \textit{F-measure} score achieved for a training set of size $[2, ~20, ~200, ~2000]$. We observe that the performance of all three algorithms under study depends on the training set. However, for a very small size of the training set, our algorithm seems to outperform the others. In that regime, the recovery performance from  the diffusion filters \cite{SegarraMMR16}  depends on the estimation of the spectral templates, which is highly dependent on the number of the training samples.  Although this approach is quite interesting and works very well when the training set is large and the estimation of the covariance matrix is accurate, it might face some limitations when the training set is limited and noisy. In contrary, our algorithm learns a graph diffusion process without making any assumption on  the eigenvectors of the graph process: it rather sets an explicit assumption on the (heat) diffusion process and the signal model. Moreover, our sparsity assumption  imposes additional structure to the problem, leading to high recovery performance even when the training set is limited.  

\begin{figure*}[t]
      \centering
            \subfigure[]
            { \includegraphics[width=6.5cm]{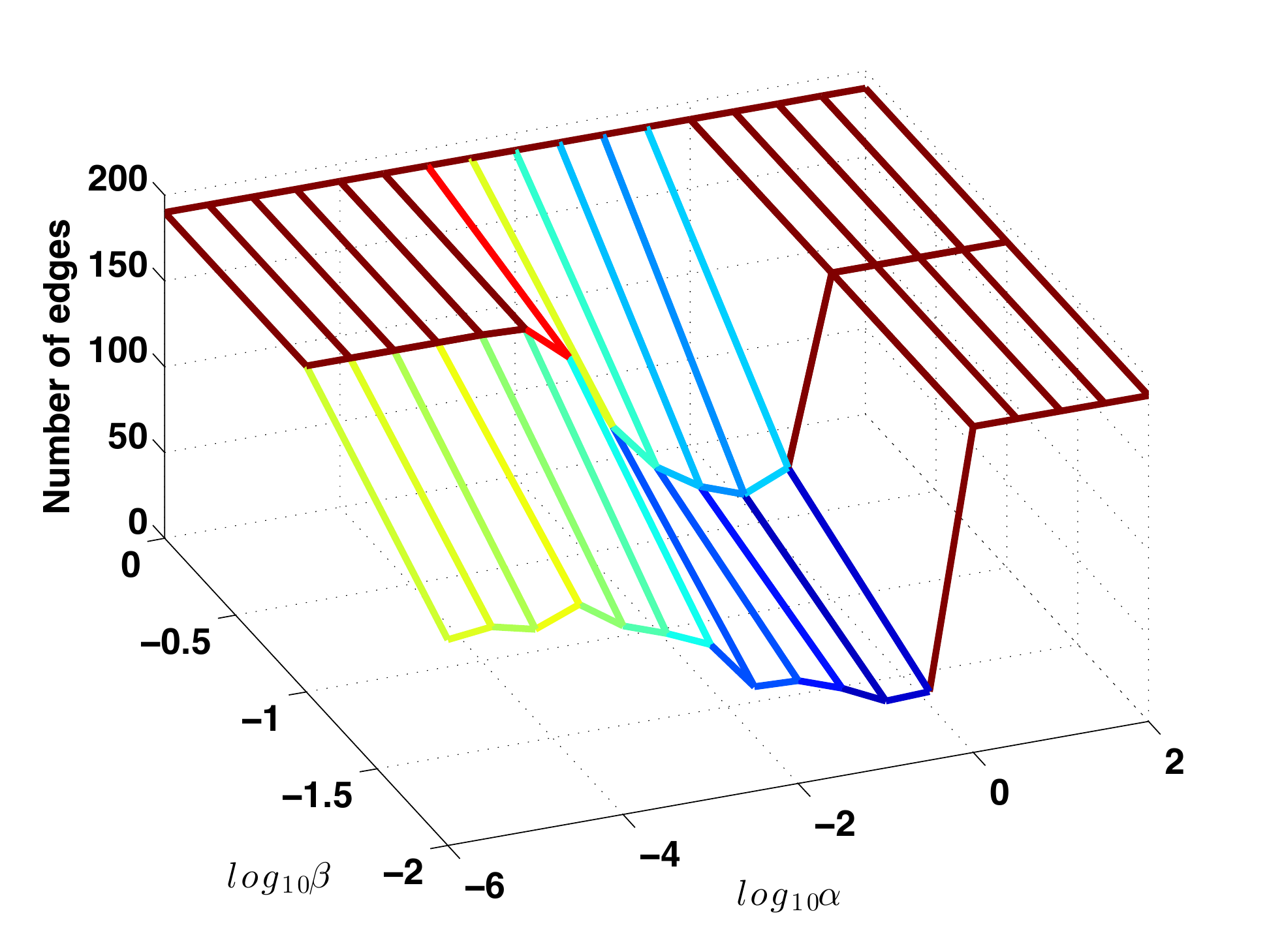}
\label{num_edges_trend_heat_kernel}}
            ~
            \subfigure[]
            { \includegraphics[width=6.5cm]{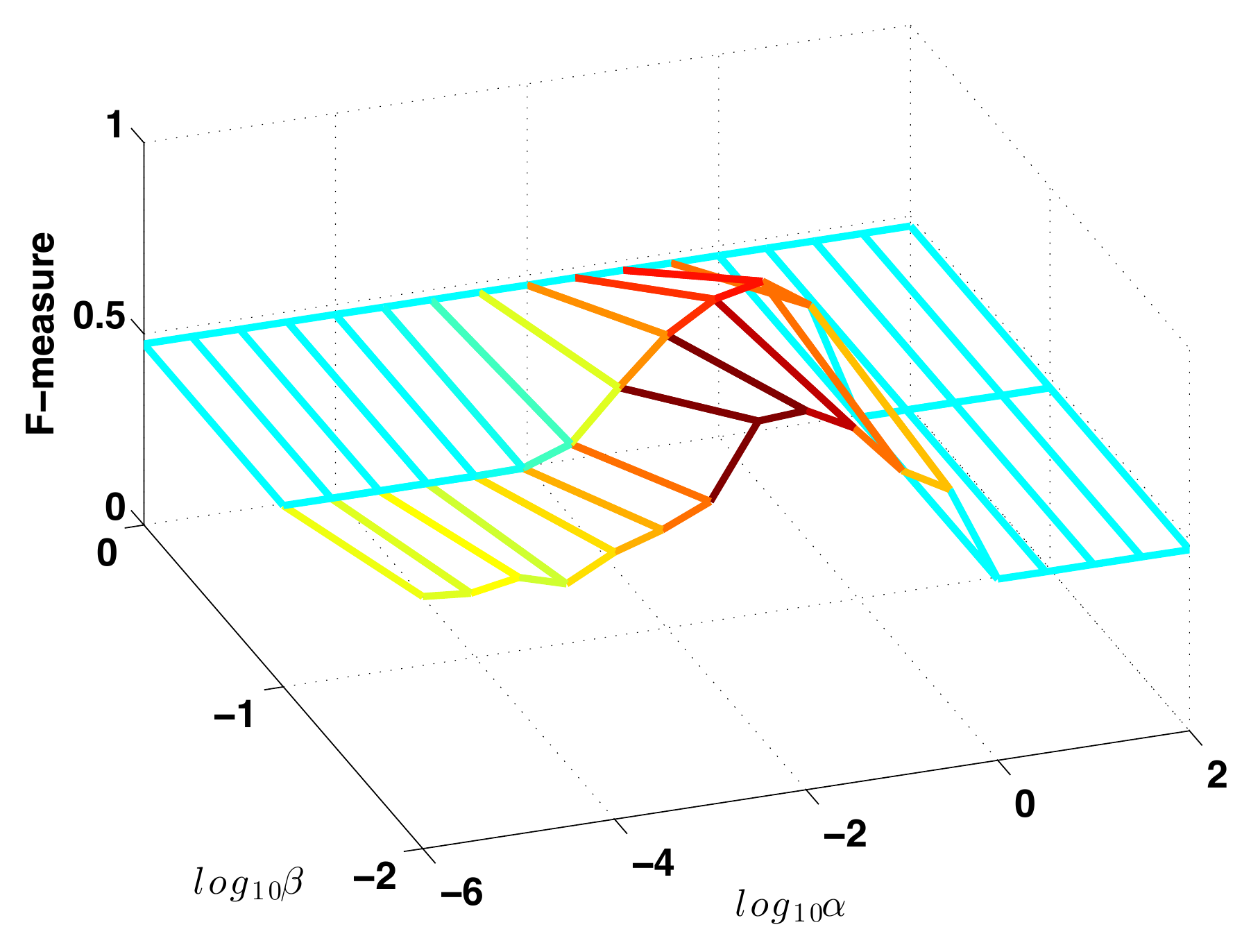}
\label{f_trend_heat_kernel}}
	\vspace{-0.2cm}
        \caption{(a) The number of edges in the learned graph, and (b) the \textit{F-measure} score, under different combinations of the parameters $\alpha$ and $\beta$ for an instance of the Gaussian RBF graph.}
        \label{fig:syn_sigrep_trend_heat_kernel}
\end{figure*}

We now study the effect of the parameters $\alpha$ and $\beta$ in the objective function of Eq. (\ref{eq:opt_prob_heat_kernel}). We illustrate in Fig. \ref{fig:syn_sigrep_trend_heat_kernel} the number of edges of the learned graph and the \textit{F-measure} score under different combinations of these parameters, for a random instance of the Gaussian RBF graph.  The obtained results indicate that the performance of the algorithm in terms of  both the number of edges and the \textit{F-measure} is mainly determined by the sparsity control parameter $\alpha$. The parameter $\beta$ influences the sparsity pattern of the learned graph. In particular, for a fixed $\alpha$, the number of learned edges decreases as $\beta$ decreases. A  big  $\beta$ implies a small Frobenius norm, which leads to a Laplacian matrix with many non-zero entries that are similar to each other.  Thus, the correct value of $\beta$ is determined by the true sparsity of the underlying graph. Then, in order to understand the effect of the parameter $\alpha$, we need to distinguish the following three cases.  When $\alpha$ is relatively small, the number of edges is relatively big, leading to a \textit{F-measure} score that is low.  In this case, the solution of the optimization problem is mainly determined by the fitting term $|| X - \D H ||^{2}_{F} $ and the Frobenius norm  constraint $ \beta \|L\|_F^2$ leading to a graph that is dense, with similar entries when $\beta$ is larger. As we increase $\alpha$, we  observe in Fig. \ref{fig:syn_sigrep_trend_heat_kernel} that the number of edges decreases, and the learned graph becomes similar to the groundtruth one as indicated by the \textit{F-measure} score. In particular, there exists a range of values for the parameter $\alpha$ where the learned graph reaches a number of edges that is similar to the one of the true graph, and the \textit{F-measure} reaches its peak. 
Alternatively, when the value of $\alpha$ is relatively big, the solution of the sparse coding step tends to give a matrix $H$ that is sparser.  In that case, the algorithm tries to express the signals in the dense matrix $X$ as a heat diffusion process starting from some sparse initial heat sources $H$. We recall that the heat kernel can be written as the Taylor expansion of the exponential function
$e^{-\tau L} = \sum_{k = 0}^{\infty}(-\tau)^k\frac{L^k}{k!}$.
Moreover, the $k^{th}$ power of the Laplacian is localized in the $k$-hop neighborhood of a node $n$, i.e., $(L^k)_{n,m} = 0$ if nodes $n$ and $m$ are not connected with a path of at least $k$-hops on the graph \cite{Hammond11}.  
 Thus, the initial heat $h$, corresponding to an observation $x$, diffuses all over the graph only if there exists a finite path connecting the sources indicated in $h$ with the other nodes of the graph. As a result,   in order to approximate a dense observation $x$,  the graph that we learn should be more connected.
In the extreme case when  $H$ is a zero matrix, the objective function penalizes only the Frobenius norm of $L$.  
The latter explains the tendency of the algorithm to favor complete graphs with similar entries when $\alpha$ is large.

\subsubsection{Source localization}

In a final set of experiments, we illustrate the performance of the learned diffusion dictionary in terms of source localization. For the sake of simplicity, we focus on the case of only one dictionary block. In particular,  we use the different instances of the learned topologies with our scheme for an RBF Gaussian graph model. We then use the learned graphs to solve a sparse inverse problem, similar to (\ref{eq:map}), to recover the sources from a set of some other signal observations. For each value of the parameter $\tau =[10^{-1} : 10^{0.5} : 10^{1.5}]$, we generate one diffusion dictionary per topology instance.  Each of these dictionaries are used to generate  a set of 1000 testing signals that are each a linear combination of 3 atoms of the corresponding dictionary, generated in the same way as the training signals. The location of these atoms defines the initial sources of the process. We aim at recovering the initial sources by solving an iterative soft thresholding algorithm \cite{Beck2009} with the diffusion dictionary on a learned graph.  

\begin{figure*}
      \centering
            \subfigure[]
            { \includegraphics[width=5cm]{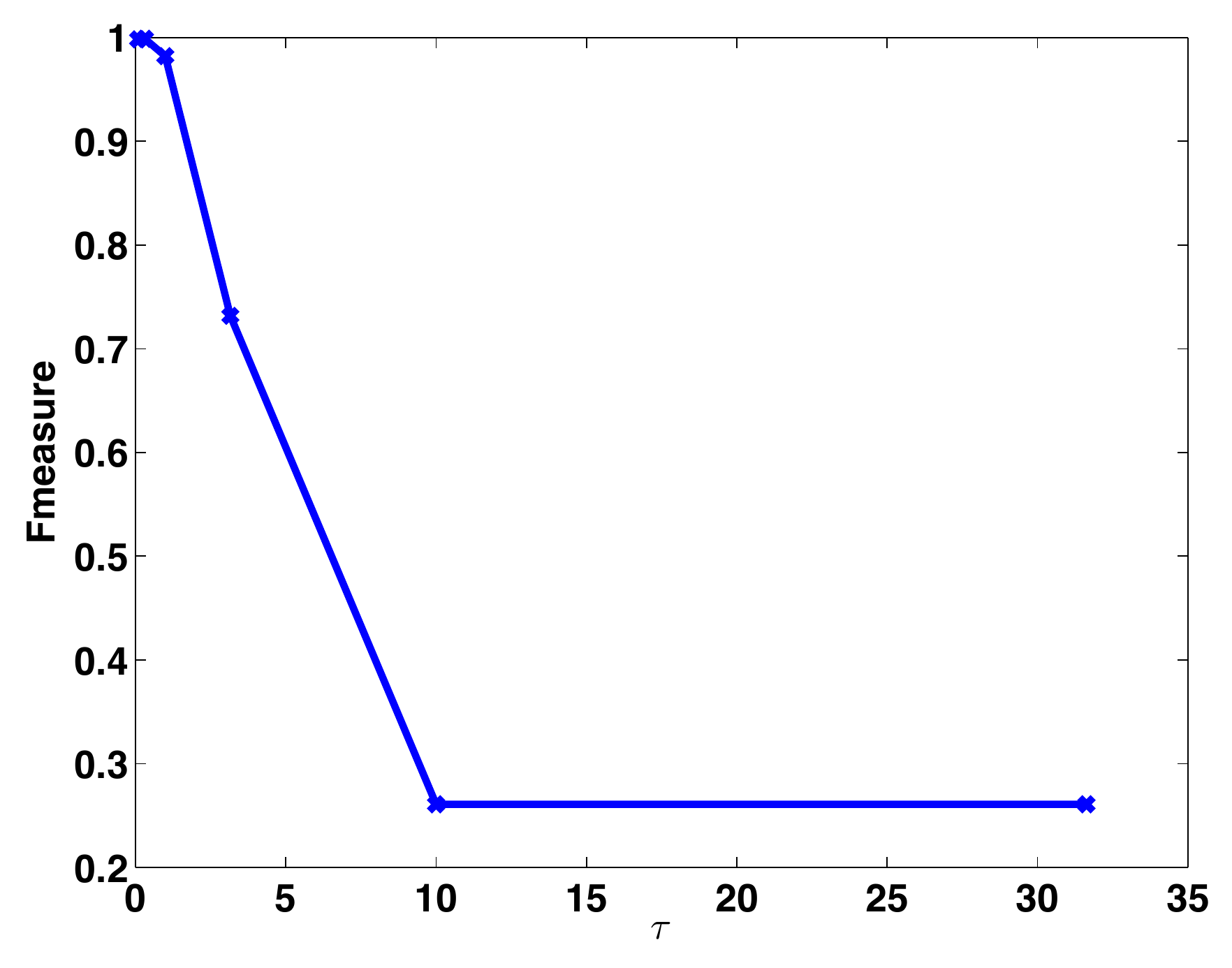}
\label{f_trend_localization}}
            ~
            \subfigure[]
            { \includegraphics[width=5cm]{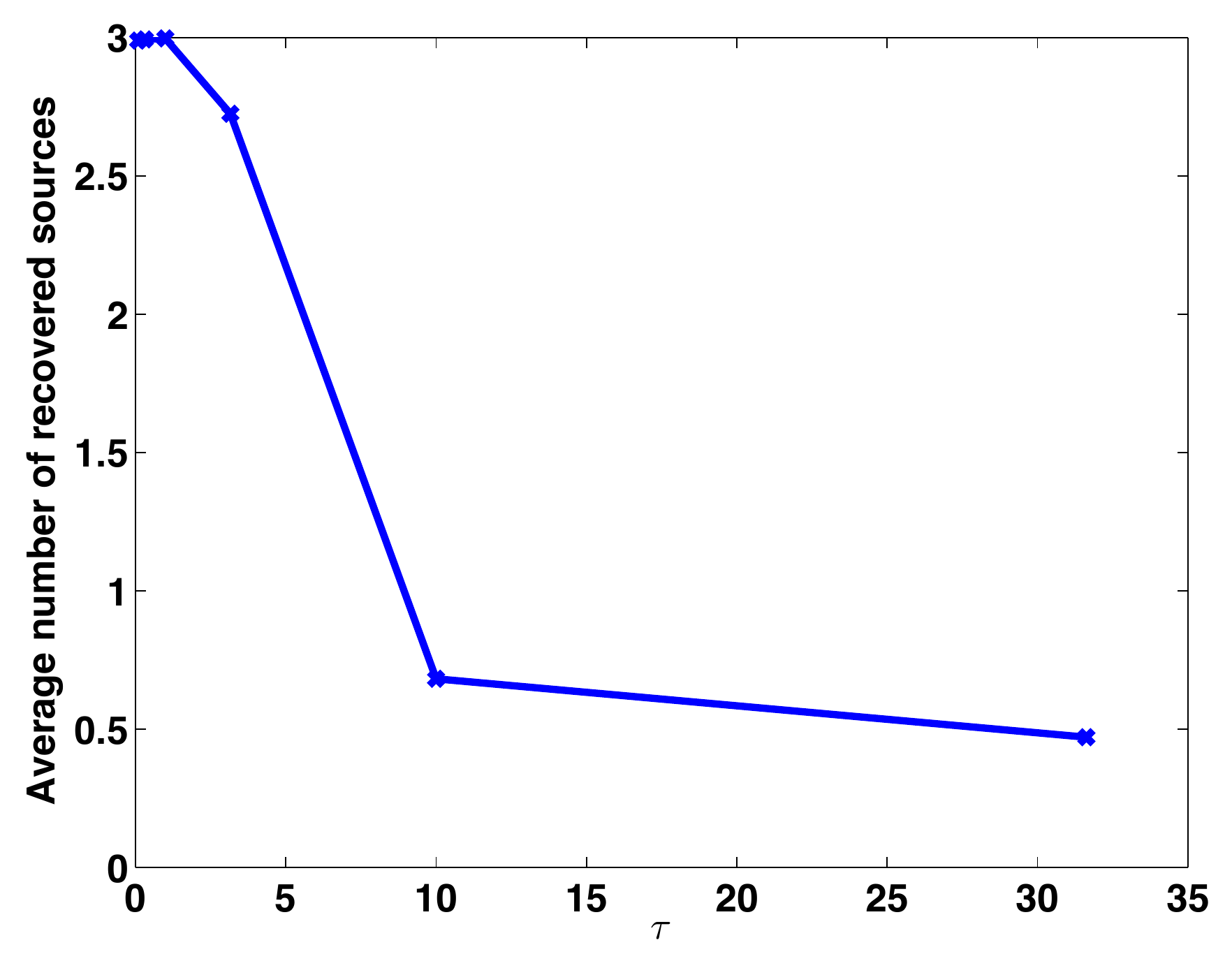}
\label{average_localization}}
	\vspace{-0.2cm}
        \caption{Source recovery performance measured with  respect to (a)   the \textit{F-measure} score between the recovered and the groundtruth sparse codes    and (b) the location of the highest in magnitude  sparse codes coefficients for different values of the diffusion parameter $\tau$, and three initial sources.}
        \label{fig:syn_localization}
\end{figure*}

In Fig. \ref{fig:syn_localization}, we show the source recovery performance for different values of the parameter $\tau$.  In particular, in Fig. \ref{f_trend_localization}, we illustrate the average  \textit{F-measure} score  between the groundtrouth sparse codes of the testing signals and the recovered ones as a function of $\tau$. We observe that the \textit{F-measure} is high when $\tau$ is low. The latter is intuitive as a small $\tau$ implies that the diffusion process is quite localized around the initial sources, leading to an easier recovery. As $\tau$ increases the performance reduces significantly, as the diffusion process tends towards a smooth signal on the graph. Thus, recovering the sources becomes more difficult.  We notice that the recovery performance is measured in terms of the activation of the sources, i.e.,   the non-zero position of the learned sparse codes, and not the actual value. In order to understand better the source localization performance with a sparse prior, we keep only the  $s$ highest ones in terms of  magnitude values of the sparse codes, where $s$ is the number of the initial sources.  We then illustrate in Fig. \ref{average_localization} the average number of recovered sources for different values of the parameter $\tau$, for the sparsity parameter $\alpha$ of (\ref{eq:map}) that gives the best source recovery results. These results are consistent with the previous ones and they confirm that 
when $\tau$ is low, the location of the sparse codes with higher magnitude refers to the initial sources.

\subsection{Graph learning in real-world datasets}
\subsubsection{ETEX dataset}
We now illustrate the performance of our algorithm on real world datasets. We first consider data from the European Tracer Experiment (ETEX), which took place in 1994 \cite{Nodop19984095}.\footnote{The dataset is publicly available in {https://rem.jrc.ec.europa.eu/etex/} and has already been processed in \cite{PenaBV16}.}   The experiment consists in injecting a particular gas, namely the tracer, into the atmospheric system and then in observing the evolution of the tracer with a variety of sampling and analysis tools. Tracer techniques are widely applied for the determination of dispersion and dilution patterns of atmospheric pollutants. In particular, an easily identifiable tracer (perfluorocarbons) has been released in the atmosphere from the city of Rennes, in France. The concentration of the tracer has then been measured over a period of 72 consecutive hours, at 168 ground-level stations in Western and Eastern Europe. In our experiments, we consider the 168 sampling stations as nodes of the graph and the concentration measured in each of them as signals on the graph. The measurements obtained at different time instances within the 72-hour period form 30 observations, which are used to infer the diffusion topology that can explain well the diffusion of the tracer. For this experiment, we choose $S = 1$ as the observations consist of many zeros entries, which indicates that they can be approximated with a single diffusion process at small scale. Moreover, we fix the scale parameter to $\tau = 3$ and we initialize the Laplacian matrix, with a random graph Laplacian matrix.
\begin{figure*}[!t]
      \centering
            \subfigure[]
{ \includegraphics[width=4.5cm]{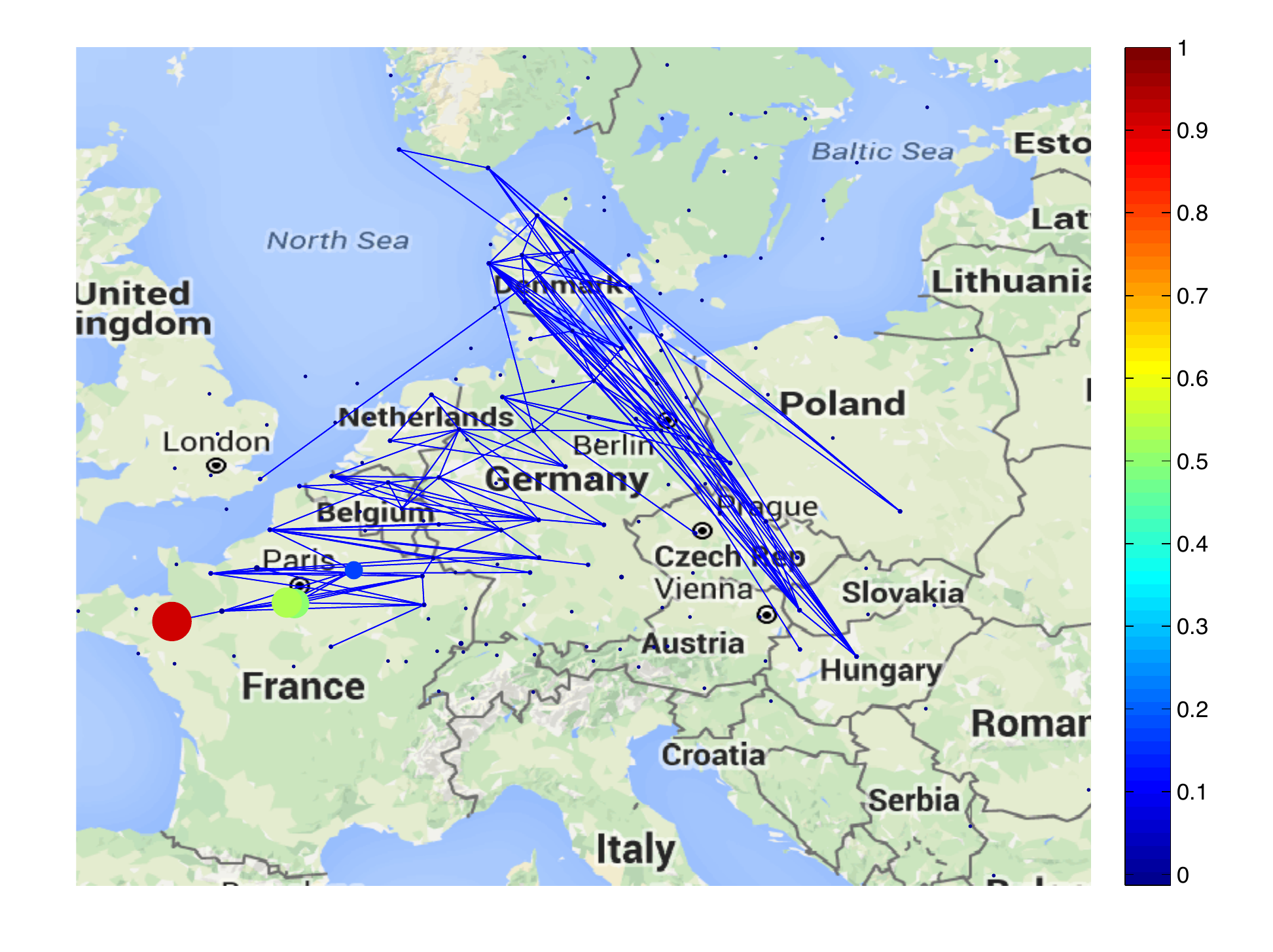}   
\label{laplacian_gt_rbf}}
            \subfigure[]
{ \includegraphics[width=4.5cm]{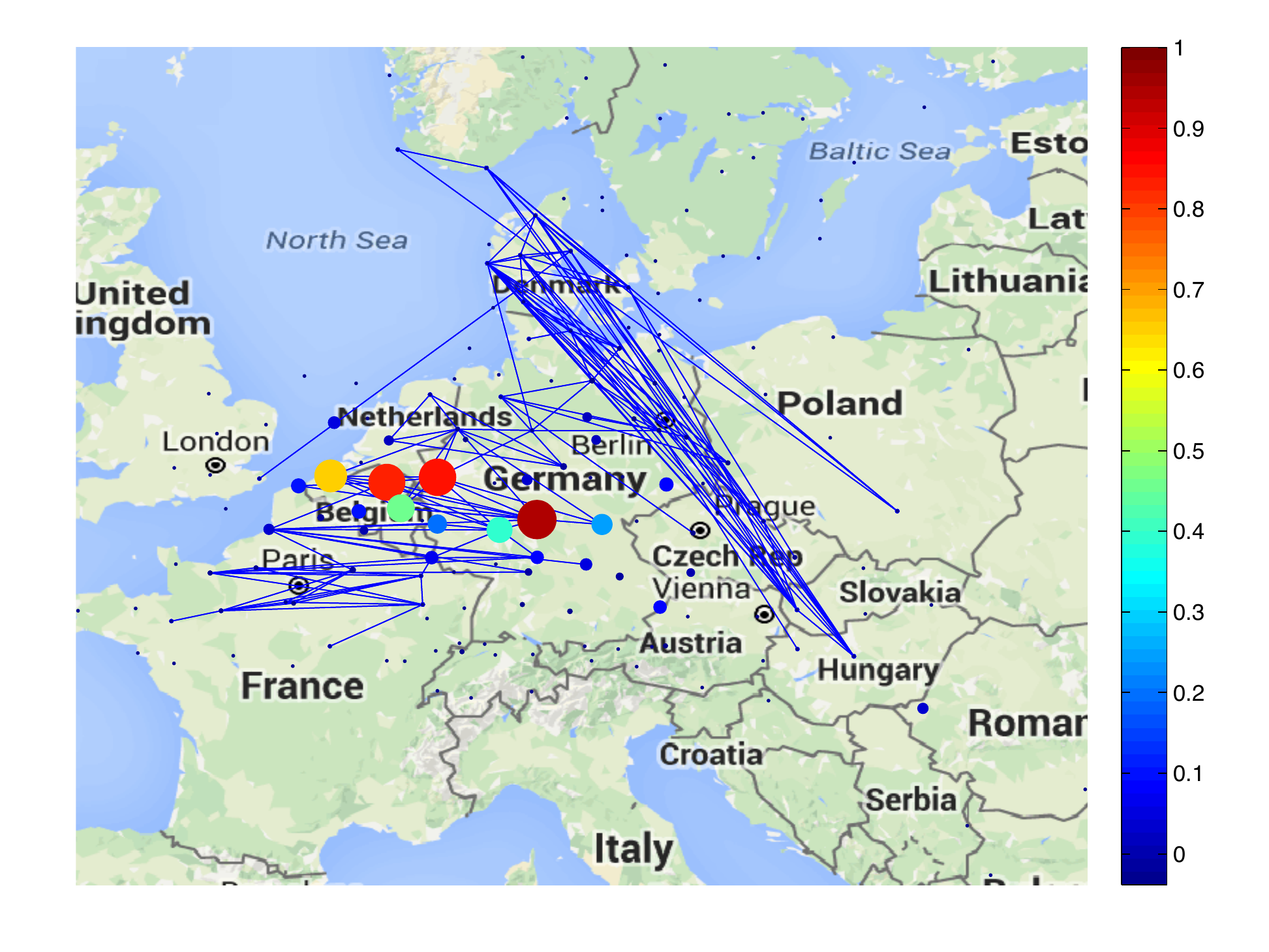}
\label{laplacian_gt_pa}}
            \subfigure[]
                { \includegraphics[width=4.5cm]{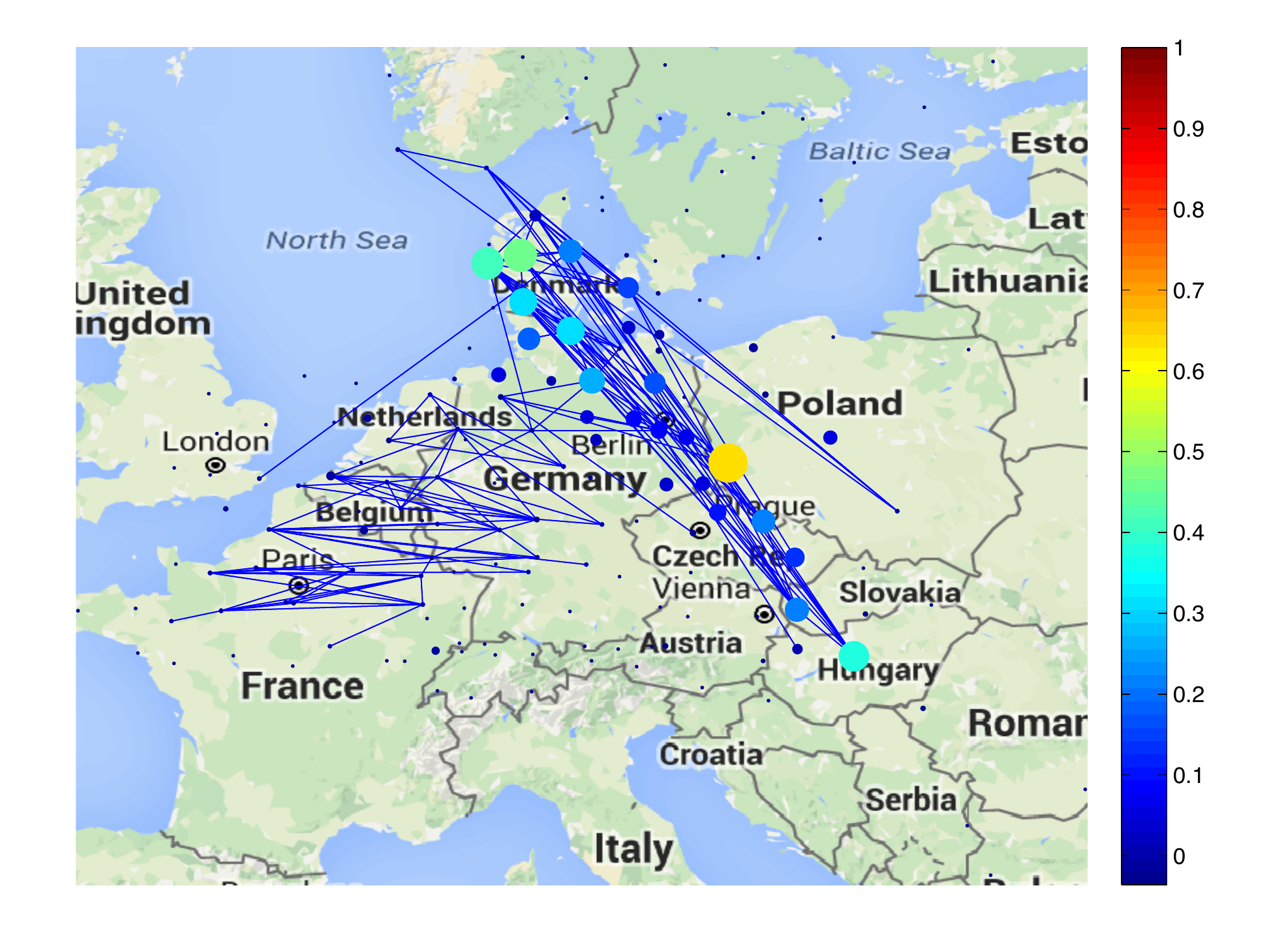}
\label{laplacian_gaussian_rbf}}
	\vspace{-0.2cm}
        \caption{The learned graph and different measurements over time (a-c) of the concentration of the tracer (signal observations). The color code represents the concentration measured in each station.}
        \label{fig:etex_examples}
\end{figure*}

In Fig. \ref{fig:etex_examples}, we illustrate the most important edges of the graph  learned with LearnHeat and some representative measurements of the concentration of the tracers, which are used as training signals in the learning. The estimated graph indicates the main directions towards which the tracer moved, which are consistent with the signal observations. These directions are influenced by many parameters such as the meteorological conditions and the direction of the wind. We observe that there exist some strong connections between stations in France and Germany, and those in Sweden and Hungary,  which are consistent with the conclusions in the documentation of the dataset \cite{Nodop19984095}.

\begin{figure}
      \centering
            { \includegraphics[width=5.5cm]{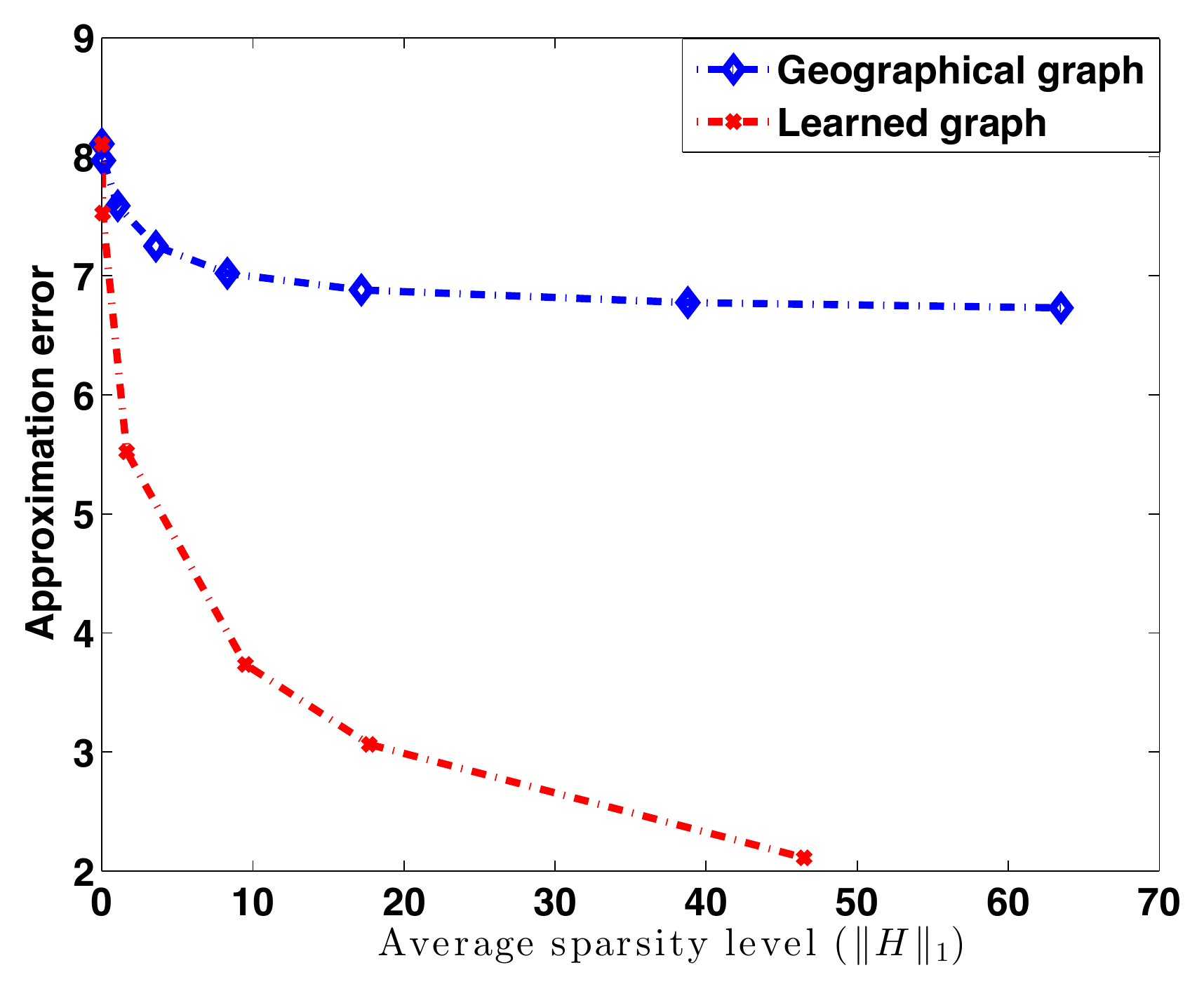}
\label{etex_sparsity}}
	\vspace{-0.3cm}
        \caption{Approximation performance of the daily signals for different sparsity levels on dictionaries generated from a geographical graph (blue) and the learned graph (red).}
        \label{fig:etex_sparsity}
\end{figure}

Finally, in Fig.~\ref{fig:etex_sparsity}, we study how well a diffusion dictionary based on the graph Laplacian can represent the signal observations with only a few atoms of the dictionary. We compute the sparse approximation using an iterative soft thresholding algorithm that promotes sparsity of the coefficients $H$. We compare our results with a diffusion dictionary designed based on a graph that is constructed using geographical distances between these stations. We observe that the approximation error $\|X-e^{-\tau L}H\|_F^2$ is significantly smaller in the case of the diffusion dictionary based on the learned graph for different sparsity levels. These results indicate that learning the topology can bring significant benefits for effective structured data representation. 

\subsubsection{Uber dataset}
\begin{figure*}[!t]
      \centering
       \subfigure[]
{ \includegraphics[width=4cm]{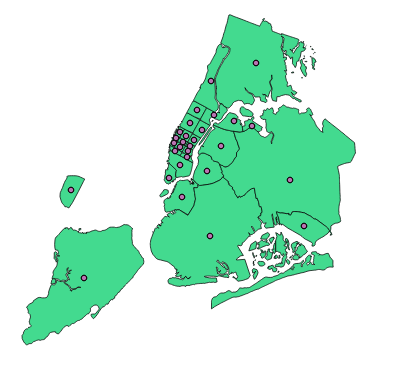}   
\label{fig:zones}}
          \subfigure[]
{ \includegraphics[width=4.5cm]{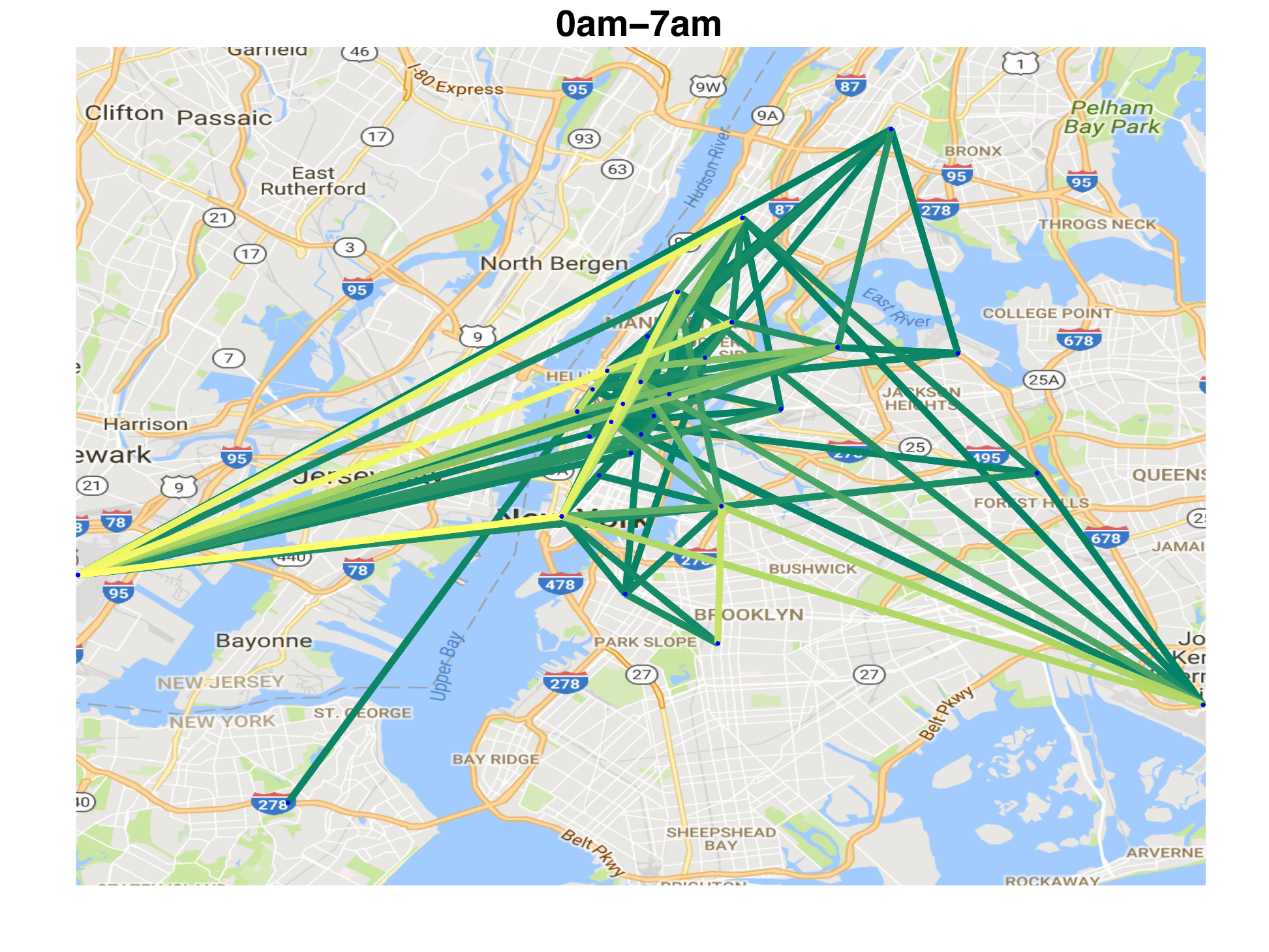}   
\label{fig:0_7}}
            \subfigure[]
{ \includegraphics[width=4.5cm]{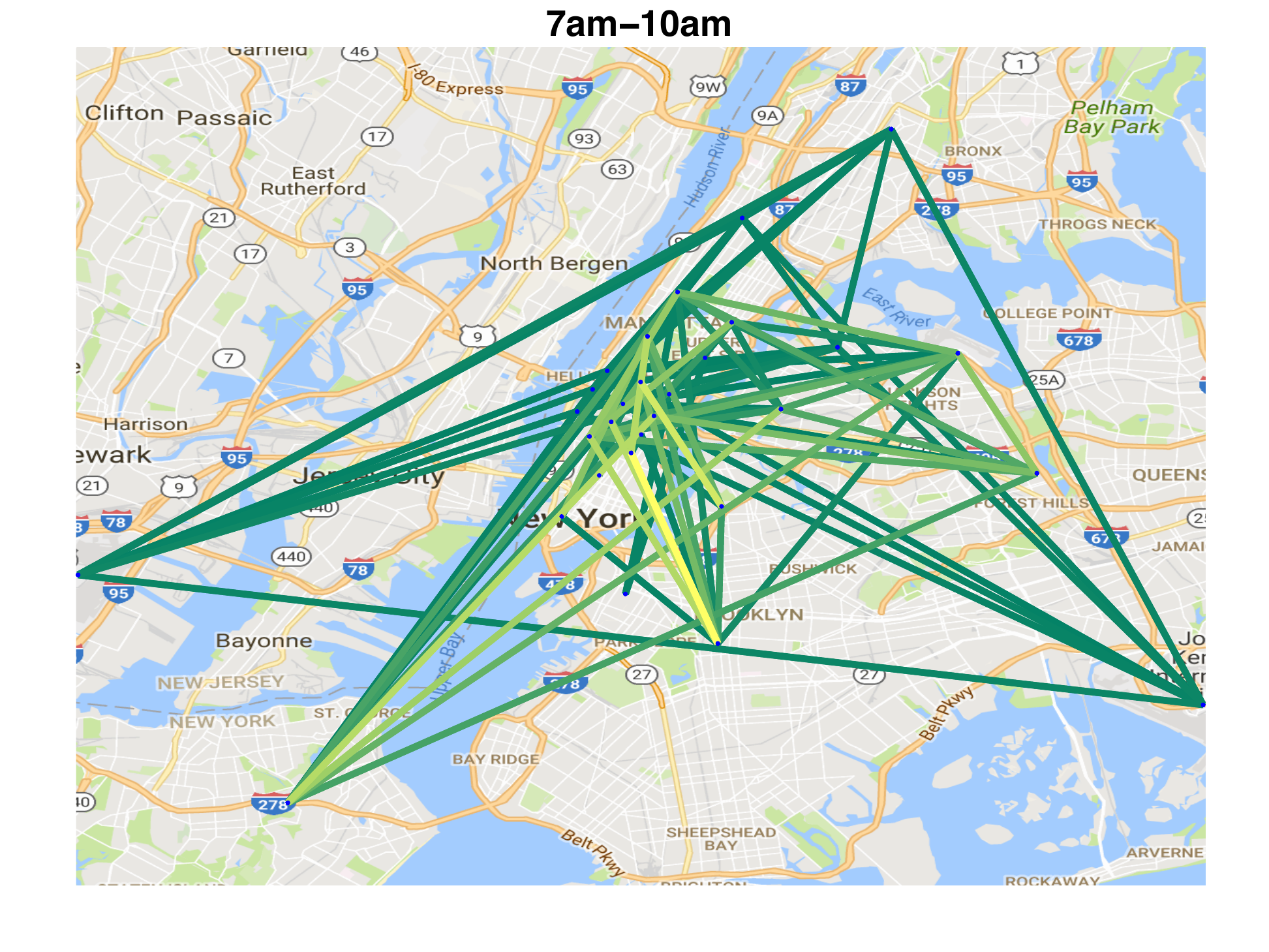}   
\label{fig:7_10}}
            \subfigure[]
{ \includegraphics[width=4.5cm]{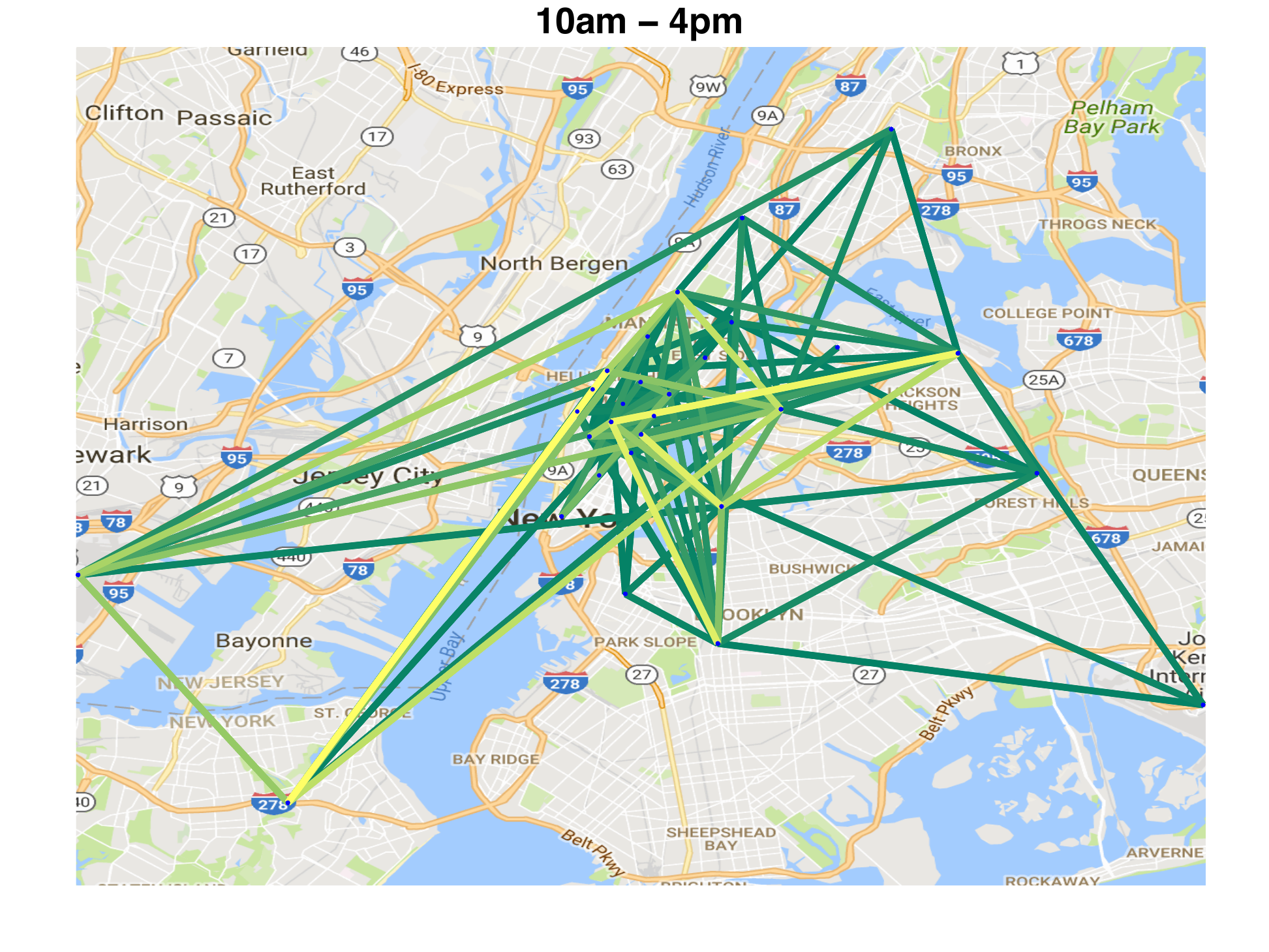}
\label{fig:10_16}}
            \subfigure[]
                { \includegraphics[width=4.5cm]{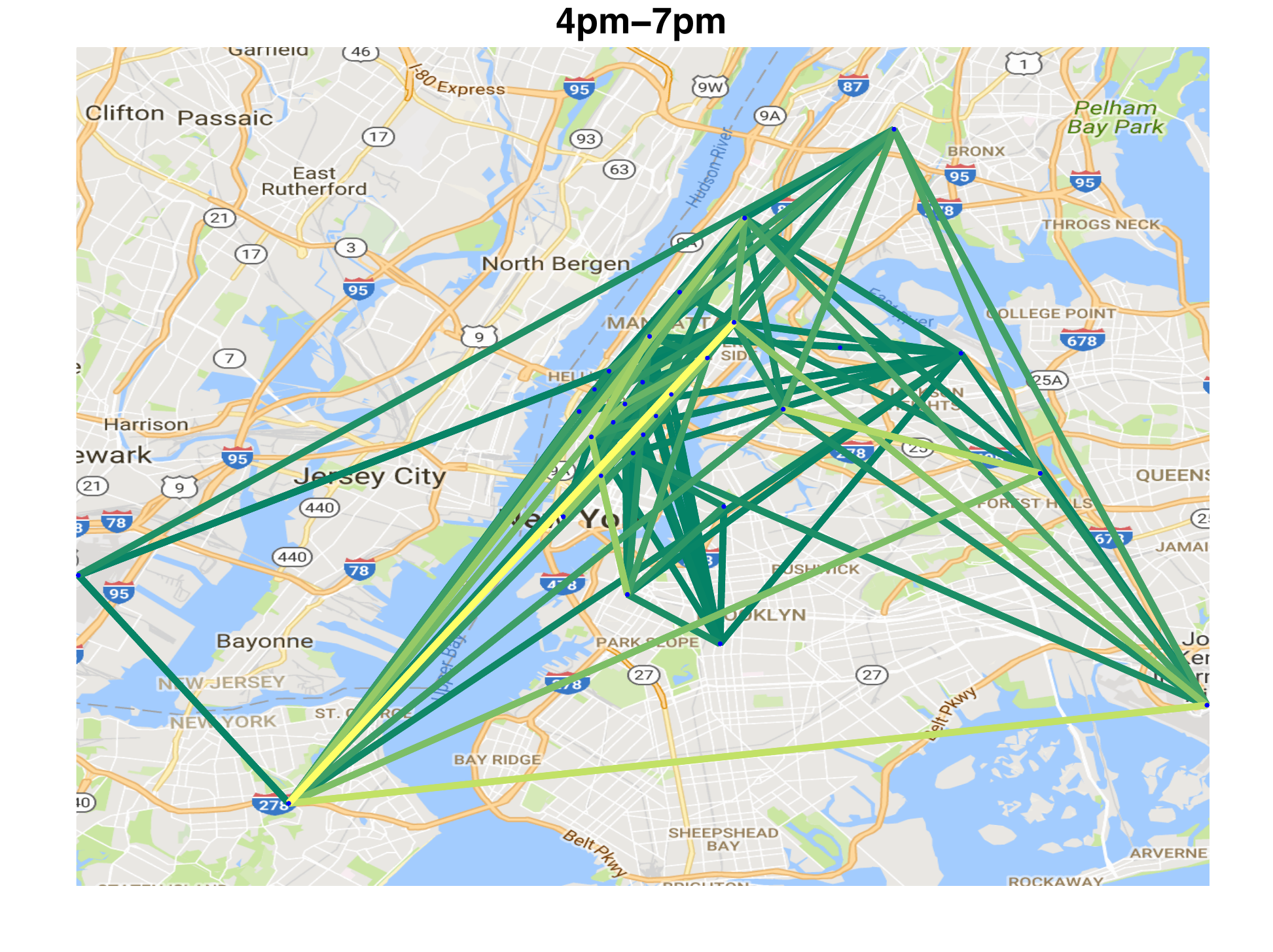}
\label{fig:16_19}}
            \subfigure[]
{ \includegraphics[width=4.5cm]{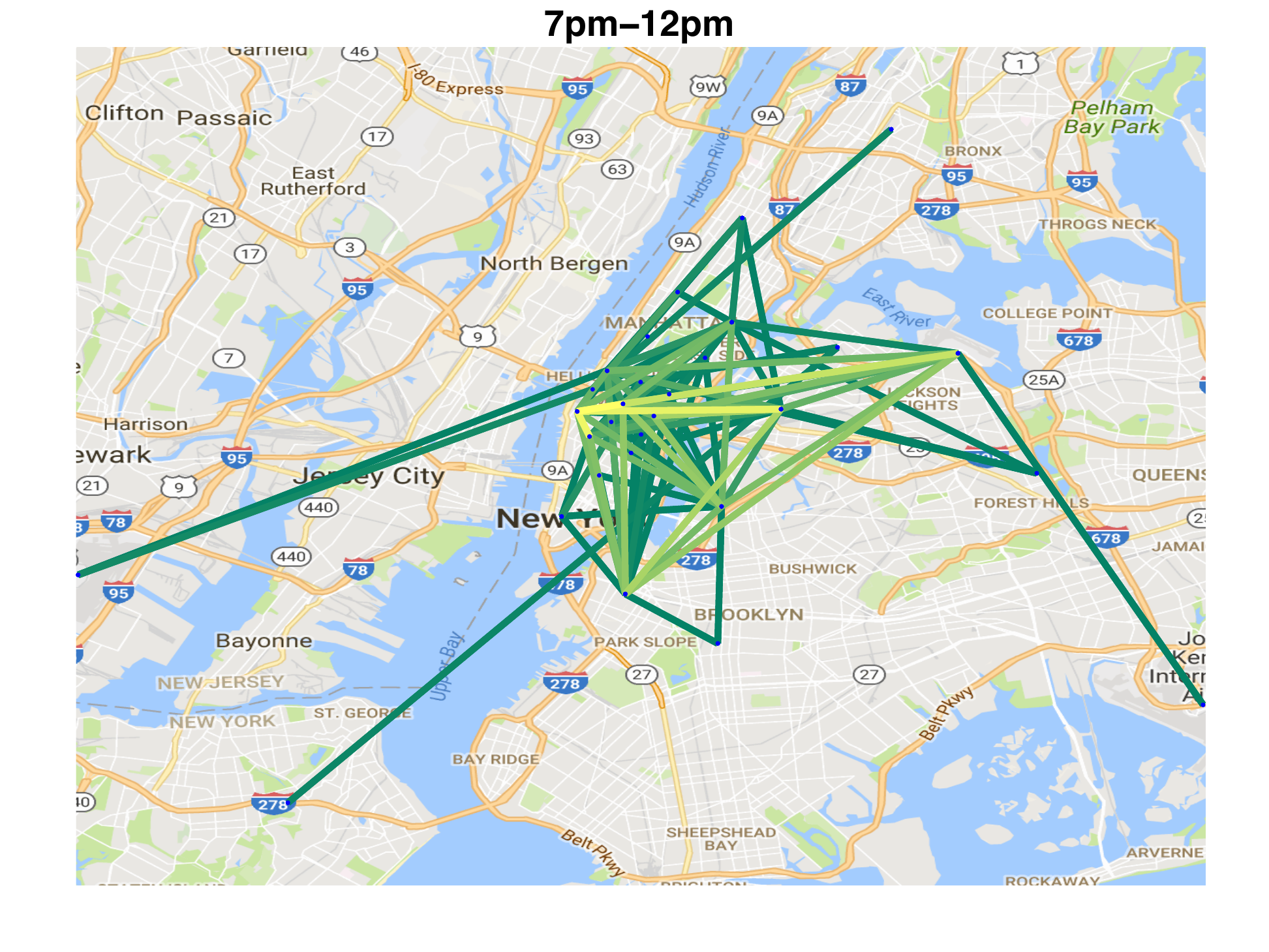}   
\label{fig:19_24}}
	\vspace{-0.2cm}
        \caption{(a) Boundaries of the taxi zones in New York City. Each zone is represented by a node on a the learned graph. The learned graphs over different time intervals: (b) 0.00 - 7.00 am, (c) 7.00 am - 10.00 am, (d) 10.00 am - 4.00 pm, (e) 4.00 pm - 7.00 pm, and (f) 7.00 pm - 12 pm.}
        \label{fig:uber_graphs}
\end{figure*}

In the final set of experiments, we use our graph learning algorithm to detect patterns from  Uber rides in New York City. In particular,  we use the Uber dataset\footnote{The dataset is publicly available in {https://github.com/fivethirtyeight/uber-tlc-foil-response}} for the month of September 2014, which provides time and location for pickups. For the sake of simplicity, we divide the city into $N = 29$ taxi zones, as shown in Fig. \ref{fig:zones}, and each zone is a node of a graph. The hourly number of Uber pickups in each zone  is a signal on the graph. Moreover,  we divide the day into five time slots 1) 7 am - 10 am, 2) 10 am - 4 pm, 3) 4 pm - 7 pm, 4) 7 pm - 12 pm, 5) 12 pm - 7 am.  For each of these slots, we define as training signals the number of pickups measured for each hour inside the corresponding time interval, all  weekdays of the month. 

For each of these five set of training signals,  we learn a heat diffusion dictionary with $S  = 2$ blocks, for different parameters of $\alpha$ and $\beta$.  In order to choose the best parameter of $\alpha$ and $\beta$, we define as a criteria that the number of edges of the learned graph should be approximately $4N$. We expect that the graph learned for each time interval conveys some information about the traffic patterns in the city.  In Fig. \ref{fig:uber_graphs}, we show the learned graphs for each of the time intervals.  We can clearly see patterns that are indicative of the behavior of the people in the city. First, there is a clear tendency of people using Uber to go/come mostly to/from airports (JFK, La Guardia, Newark)  in early morning (Fig. \ref{fig:0_7}). Moreover,  the connections of the graph during the rush hours (7 am - 10 am and 4 pm - 7 pm) indicate the commuting of people from/to different neighborhood of the city to/from Manhattan. During the day (10 am - 4 pm), there is no clear pattern as the graph learned from the distribution of the Uber cars indicates that people tend to use Uber to go to random places in the city. Finally, from 7 pm to midnight (Fig. \ref{fig:19_24}), most of the connections are concentrated across Manhattan, which probably indicates that most of the people use Uber to visit bars or restaurants that are concentrated around that area.  These are of course just some observations that confirm the efficiency of our algorithm in learning meaningful graphs. A more detailed mining of the mobility patterns in New York City requires taking into consideration other factors such as the population of each region, a finer grid of the zone, the organization of the city in terms of public transform, which is out of the scope of this paper.

\section{Conclusion}

In this paper, we have presented a framework for learning graph topologies (graph Laplacians) from signal observations under the assumption that the signals are generated from  heat diffusion processes  starting from a few nodes of the graph. Specifically, we have proposed an algorithm for learning graphs that enforces sparsity  of the graph signals in a dictionary that is  a concatenation of graph diffusion kernels at different scales. Experimental results on both synthetic and real world diffusion processes have confirmed the usefulness of the proposed algorithm in recovering a meaningful graph topology and thus leading to better data understanding and inference. We believe that the proposed graph learning framework  opens new perspectives in the field of data inference and information propagation in particular from the emerging graph signal processing viewpoint. %



%

\section{\textsc{Acknowledgments}}
The authors would like to thank S. Segarra and A. Ribeiro for sharing their MATLAB code of the algorithm in \cite{SegarraMMR16} used in the experiments, and G. Stathopoulos for discussions on the solution of the optimization problem.  

\appendices

\section{Computation of the gradient}
\label{appendixA}
As noted in Section \ref{sec:graph_learning}, Algorithm \ref{Alg_Heat_Kernel} require the computation of the gradient of the fitting term with respect to each of the variables $H, L, \tau$. In the following, we discuss the computation of each of the gradients separately.
\subsection{Gradient with respect to $H$}
\label{appendixA_H}
The gradient of $Z(L, H, \tau) = \|X -\D H\|_{F}^2$ with respect a column $h_j$ of  $H$ is independent of the other columns of $H$. Moreover, it depends only on the corresponding observation $x_j$ and it can be written as
\begin{equation}
\nabla Z_{h_j}(L^{t}, H^t, \tau^t) = -2\D^T (x_j -\D  h_j).
\end{equation}

\subsection{Gradient with respect to $L$} 
\label{appendixA_L}
The gradient of $ \|X -\D H\|_{F}^2$ with respect to $L$ is:
\begin{align}
\nonumber &\nabla_L \|X -\D H\|_{F}^2\\ 
\nonumber & = \nabla_L \big(\tr \big((X -\sum_{s=1}^{S}e^{-\tau_s L} H_s)^T(X -\sum_{s=1}^{S}e^{-\tau_s L} H_s)\big)\big)\\
\nonumber & =   \nabla_L \big( \tr(X^T X) -2\sum_{s=1}^{S}\tr(H_sX^T e^{-\tau_s L}) \\ \nonumber &\quad \quad + \sum_{s=1}^{S}\sum_{s'=1}^{S}\tr(H_{s'}H_s^T e^{-(\tau_s + \tau_{s'}) L})\big)\\
\label{eq:gradientL}& =   -2 \sum_{s=1}^{S}\nabla_L\tr(H_sX^T e^{-\tau_s L}) \\\nonumber  &\quad \quad  +  \sum_{s=1}^{S}\sum_{s'=1}^{S}\nabla_L\tr(H_{s'}H_s^T e^{-(\tau_s + \tau_{s'}) L}).
\end{align}
In order to compute Eq. \eqref{eq:gradientL}, we make use of the following proposition. 

 \begin{proposition}  \label{proposition}
  Consider a general matrix $A\in \mathbb{R}^{N\times N}$ and a symmetric matrix $L \in \mathbb{R}^{N\times N}$, admitting  a spectral decomposition $L = \chi \Lambda \chi^T$. Then 
  \begin{equation*}
  \nabla_L \tr(A e^{L}) = \chi \big((\chi^T A^T \chi) \circ B \big)\chi^T,
  \end{equation*}
 where $\circ$ denotes the Hadamard product and $B$ is the $N\times N$ matrix defined by the entries
\begin{equation}
\label{eq:Bdef}[B]_{ij} =  \begin{cases}
    e^{\Lambda_{ii}}      & \quad \text{if } \Lambda_{ii} = \Lambda_{jj} \\
    \frac{e^{\Lambda_{ii}}  - e^{\Lambda_{jj}} }{\Lambda_{ii} - \Lambda_{jj}} & \quad \text{otherwise.}
  \end{cases}
\end{equation}

   \begin{proof} \rm 
The desired gradient is uniquely defined by satisfying the relation 
\begin{equation}
\label{eq:gradApprox} \tr(A e^{ (L + \Delta)}) - \tr(Ae^{ L}) = \langle \nabla_L \tr(A e^{L}), \Delta \rangle + \mathcal{O}(\|\Delta\|^2)
\end{equation}
for all sufficiently small perturbations $\Delta$.
Using the fact that the eigenvectors of $L$ are orthonormal, i.e., $\chi^T \chi = I$, where $I$ is the identity matrix, we can write the left hand-side of Eq. \eqref{eq:gradApprox} as follows:
\begin{align}
\nonumber &\tr(A e^{ (L + \Delta)}) - \tr(Ae^{ L}) \\ \nonumber & = \tr(\chi^T A \chi \chi^T e^{ (L + \Delta)}\chi) - \tr(\chi^T A \chi \chi^T e^{ L}\chi) \\
& = \tr(\chi^T A \chi e^{ (\Lambda + \chi^T\Delta\chi)}) - \tr(\chi^T A \chi e^{ \Lambda}). \label{eq:decomposedDer}
\end{align}
The Fr\`echet derivative of the matrix exponential at a diagonal matrix $\Lambda$ applied to a direction $\Delta$ is the $N\times N$ matrix
$D e^{\Lambda}(\Delta) = B \circ \Delta$ with $B$ defined in~\eqref{eq:Bdef}; see~\cite{Higham:2008}.
Using the above developments and the linearity of the trace operator we obtain that  
\begin{align}
\nonumber& \langle\nabla_{\Lambda} \tr(\chi^T A \chi e^{ \Lambda)}), \Delta\rangle = \tr(\chi^T A \chi D e^{\Lambda}(\Delta)) \\
& =  \tr(\chi^T A \chi (B\circ \Delta))  =  \langle\chi^T A^T \chi \circ B, \Delta \rangle. \label{eq:DerivativeBD}
\end{align}
Finally,  using again the orthonormality of the eigenvectors $\chi$, we can write 
\begin{align}
\nonumber\langle \nabla_L \tr(A e^{ L}), \Delta\rangle& = \langle \chi^T\nabla_L \tr(A e^{ L}) \chi, \chi^T\Delta\chi \rangle\\
& \overset{\eqref{eq:decomposedDer}}= \langle\nabla_{\Lambda}\tr(\chi^TA\chi e^{- \Lambda}) , \chi^T\Delta\chi \rangle \nonumber \\
& = \langle\chi \nabla_{\Lambda}\tr(\chi^TA\chi e^{- \Lambda}) \chi^T, \Delta\rangle. \label{eq:decomposedDer2}
\end{align}
Combining   Eqs. \eqref{eq:DerivativeBD}, \eqref{eq:decomposedDer2}, we conclude that $\nabla_L \tr(A e^{ L}) = \chi(\chi^T A^T \chi \circ B) \chi^T$.   \end{proof}
  \end{proposition}

  Given the result of Proposition~\ref{proposition}, the gradient  $\nabla_L \tr(A e^{\nu L})$ for some $\nu \in \mathbb{R}$ can be found by applying the chain rule: $ \nabla_L \tr(A e^{\nu L}) =\nu  \nabla_{\nu L} \tr(A e^{\nu L})$. 
    
  \subsection{Gradient with respect to $\tau$}
  \label{appendixA_tau}
  
The gradient of $ \|X -\D H\|_{F}^2$ with respect to $\tau$ satisfies
\begin{align}
\nonumber &\nabla_\tau \|X -\D H\|_{F}^2\\ 
\nonumber & = \nabla_\tau \big(\tr \big((X -\sum_{s=1}^{S}e^{-\tau_s L} H_s)^T(X -\sum_{s=1}^{S}e^{-\tau_s L} H_s)\big)\big)\\
\nonumber &\quad \quad + \sum_{s=1}^{S}\sum_{s'=1}^{S}\tr(H_{s'}H_s^T e^{-(\tau_s + \tau_{s'}) L})\big)\\
\label{eq:gradient_tau}& =   -2 \sum_{s=1}^{S}\nabla_\tau\tr(H_sX^T e^{-\tau_s L}) \\\nonumber  &\quad \quad  +  \sum_{s=1}^{S}\sum_{s'=1}^{S}\nabla_\tau\tr(H_{s'}H_s^T e^{-(\tau_s + \tau_{s'}) L}).
\end{align}
By the Taylor expansion of the exponential, it follows for any $A\in \mathbb{R}^{N\times N}$ that
\begin{equation}
\label{tau_gradient}\nabla_{\tau_s}\tr(Ae^{-\tau_s L}) = -\tr(A L e^{-\tau_s L}).
\end{equation} 
Combining (\refeq{eq:gradient_tau}), (\refeq{tau_gradient}), we obtain that 
\begin{align}
\nonumber &\nabla_{\tau_s} \|X -\D H\|_{F}^2 =   2 \tr(H_sX^T L e^{-\tau_s L}) \\\nonumber  &\quad \quad  -  2\sum_{s'=1}^{S}\tr(H_{s'}H_s^T L e^{-(\tau_s + \tau_{s'}) L}).
\end{align}
Finally, the gradient with respect to the vector $\tau$ is given by a vector whose elements consist of the gradient with respect to each element of $\tau$, i.e., $\nabla_\tau \|X -\D H\|_{F}^2 = \big\{ \nabla_{\tau_s} \|X -\D H\|_{F}^2  \big\}_{s = 1}^S.$
  
  \section{Computation of the Lipschitz constants}
  \label{appendixB}
A condition for ensuring convergence of PALM is that at each iteration of the algorithm the descent lemma is satisfied \cite{Bolte_2014}. This, however, requires to determine a global Lipschitz constant or an approximation thereof such that the descent condition is satisfied.   Next, we discuss the computation of the Lipschitz constants related to the update of each of the  three variables $L, H, \tau$ in our  graph learning algorithm. As we will see, it is feasible to compute these constants for the update of $H$ and $\tau$. On the other hand, the computation of the Lipschitz constant more difficult for $L$ because of the involved matrix exponential. In this case, we perform backtracking to approximate the Lipschitz constant.
  \subsection{Variable $H$}
   \label{appendixB_H}
  The function $\nabla_H Z(H, L, \tau)$ is globally Lipschitz with Lipschitz constant $C_1(L, \tau) =\|2\D^T \D \|_F$, as can be seen from
  \begin{align}
 \nonumber & \|\nabla_HZ(L, H_1, \tau) - \nabla_HZ( L, H_2,\tau)\|_F  \\
\nonumber  & = \|-2\D^T (X -\D  H_1) + 2\D^T (X -\D H_2)\|_F \\
\nonumber  & = \|2\D^T \D  H_1 - 2\D^T \D H_2\|_F \\
\nonumber & \le \|2\D^T \D \|_F\|H_1 - H_2\|_F,
  \end{align}
  
  \subsection{Variable $L$}
   \label{appendixB_L}
  Due to the difficulty of computing the Lipschitz constant for an exponential matrix function, we estimate the associated constant $C_2(H, \tau)$ by performing backtracking line search as follows. One condition for convergence of PALM is that the descent lemma is satisfied at each iteration, i.e.,  
  
  \begin{align} 
  \nonumber &  Z(L^{t+1}, H^{t+1}, \tau^t) 
 \label{descent_condition}   \le  Z(L^{t}, H^{t+1}, \tau^t) \\ &+  \nabla_L Z(L^{t}, H^{t+1}, \tau^t)^T (L^{t+1} - L^{t})  + \frac{C_2(H, \tau)}{2}\|L^{t+1} - L^{t}\|_F^2.
  \end{align}
Moreover, the solution $L^{t+1}$ of the optimization problem (\refeq{eq:updateLQP}) indicates that for every $L \in \mathcal{C}$
\begin{align*}
& \langle L^{t+1} - L^t, \nabla_L Z(L^{t},  H^{t+1}, \tau^t) \rangle + \frac{d_t}{2}\| L^{t+1} - L^{t} \|^{2}_{F} + \beta \|L\|_F^2 \\
\nonumber  &  \le \langle L - L^t, \nabla_L Z(L^{t}, H^{t+1}, \tau^t) \rangle + \frac{d_t}{2}\| L - L^{t} \|^{2}_{F} + \beta \|L\|_F^2.
 \end{align*}
 By setting $L = L^t$ in the right-hand side of the inequality and combining with Eq. (\refeq{descent_condition}), we obtain that 
 \begin{equation}
 Z(L^{t+1}, H^{t+1}, \tau^t) \le  Z(L^{t+1}, H^{t+1}, \tau^t),
 \end{equation}
 where we have used the fact that $d_t \ge C_2(H, \tau)$. This result guarantees the decrease of the objective function over the iterations. 
The backtracking is shown in Algorithm \ref{Backtracking}.  
\begin{algorithm}[t]
\caption{Backtracking algorithm for estimating $C_2(H,\tau)$ at iteration $t+1$}
\begin{algorithmic}
\item [ 1:] {\bf Input}:  $\eta = 1.1$, initial guess for $C_2(H, \tau)$, $k = 1$
\item [ 2:] {\bf Output:}  Estimate of the Lipschitz constant $C_2(H, \tau)$
\item [ 3:] {{\bfseries while} (\ref{descent_condition}) is False  \bfseries{do}:} 
\item [ 4:]  \quad {Update: \small{$C_2(H, \tau) =  \eta^k C_2(H, \tau)$, $d_t  = \gamma_2 C_2(H, \tau)$}}
\item [ 5:]  \quad $k = k + 1$
\item [ 6:]  \quad {Update $L^{t+1}$ by solving Eq. (\refeq{eq:updateLQP}) }
\end{algorithmic}
\label{Backtracking}
\end{algorithm}

  \subsection{Variable $\tau$}
   \label{appendixB_tau}
%
  Since the objective function is convex and twice differentiable with respect to $\tau$, we estimate the Lipschitz $C_3(L, H)$ by computing the Hessian $\nabla_\tau^2 \|X -\D H\|_{F}^2$. Using~(\ref{eq:gradient_tau}), the entries of this $S\times S$ matrix are given by 
\begin{align}
\nonumber \nabla_\tau^2Z_{ss} =  &-2 \tr(H_sX^T L^2e^{-\tau_s L}) + 4 \tr(H_sH_s^T L^2e^{-\tau_s L}) \\
& +2  \sum_{s'\ne s=1}^{S}\tr(H_{s'}H_s^T L^2e^{-(\tau_s + \tau_{s'}) L}),
\end{align}
\begin{align}
\nonumber \nabla_\tau^2Z_{ss'} =   2  \tr(H_{s'}H_s^T L^2e^{-(\tau_s + \tau_{s'}) L}), \mbox{ if $s\ne s'$}.
\end{align}
Given that the Hessian is a positive semidefinite matrix, its $2$-norm is its largest eigenvalue and any upper bound on this eigenvalue gives a global Lipschitz constant. 
We will use the fact that the largest absolute row sum of the matrix represents such an upper bound. For this purpose, we first estimate 
  \begin{align}
\nonumber |\nabla_\tau^2Z_{ss}| \le  &\big(2 \|H_s\|_{F}\|X^T\|_{F}  + 4 \|H_s\|_{F}\|H_s^T\|_{F}\big) \|L^2\|_{2}^2 \\
\nonumber& +2  \sum_{s'\ne s=1}^{S}\|H_{s'}\|_{F}\|H_s^T\|_{F} \|L^2\|_{2}^2,
\end{align}
  \begin{align}
\nonumber |\nabla_\tau^2Z_{ss'}| \le \|H_{s'}\|_{F}\|H_s^T\|_{F} \|L^2\|_{2}^2,
\end{align}
where we have used the fact that $\|L^2e^{-\tau_s L}\|_2 \le \|L^2\|_2$, for every $\tau_s\ge0$, due to the positive semidefiniteness of $L$. An upper bound on the largest eigenvalue, which in turn gives the Lipschitz constant, is thus given by
\begin{equation*}
\small{
C_3(L, H) = \underset{s'}\max \|L\|_2^2 (2\|H_{s'}\|_F \|X\|_F + 4 \sum_{s=1}^S \|H_{s'}\|_F \|H_s\|_F).
}
\end{equation*}

\bibliographystyle{IEEEtran.bst}
\bibliography{mybibfile.bib}

\end{document}